\documentclass[journal,transmag]{IEEEtran}

\usepackage[pdfstartview=XYZ,
bookmarks=true,
colorlinks=true,
linkcolor=blue,
urlcolor=blue,
citecolor=blue,
pdftex,
bookmarks=true,
linktocpage=true, 
hyperindex=true
]{hyperref}

\usepackage{cite}   
\usepackage{hyperref}  

\usepackage{orcidlink}
\usepackage{bm}
\usepackage{amsmath}
\usepackage{amssymb}
\usepackage{multirow}
\usepackage{graphicx}
\usepackage{subcaption}
\usepackage{cleveref}
\usepackage{siunitx}
\usepackage{ mathrsfs }

\begin{document}

\title{Magnetic Hysteresis Modeling with Neural Operators}

\author{\IEEEauthorblockN{
Abhishek Chandra\textsuperscript{\orcidlink{0000-0003-2319-6221}}\IEEEauthorrefmark{1,2},~\IEEEmembership{Graduate Student Member,~IEEE}, 
Bram Daniels\textsuperscript{\orcidlink{0000-0002-0037-9327}}\IEEEauthorrefmark{1},~\IEEEmembership{Graduate Student Member,~IEEE}, \\
Mitrofan Curti\textsuperscript{\orcidlink{0000-0002-0084-4372}}\IEEEauthorrefmark{1,2},~\IEEEmembership{Member,~IEEE},
Koen Tiels\textsuperscript{\orcidlink{0000-0001-9279-110X}}\IEEEauthorrefmark{2,3},~\IEEEmembership{Member,~IEEE}, and
Elena A. Lomonova\textsuperscript{\orcidlink{0000-0002-2515-1441}}\IEEEauthorrefmark{1,2},~\IEEEmembership{Senior Member,~IEEE}}
\IEEEauthorblockA{\IEEEauthorrefmark{1}Department of Electrical Engineering, Eindhoven University of Technology, 5612 AP Eindhoven, The Netherlands}
\IEEEauthorblockA{\IEEEauthorrefmark{2}Eindhoven Artificial Intelligence Systems Institute, Eindhoven University of Technology, 5612 AP Eindhoven, The Netherlands}
\IEEEauthorblockA{\IEEEauthorrefmark{3}Department of Mechanical Engineering, Eindhoven University of Technology, 5600 MB Eindhoven, The Netherlands}
\thanks{Manuscript received July 3, 2024, revised October 14, 2024, accepted November 9, 2024. 
Corresponding author: A. Chandra (email: a.chandra@tue.nl).}}

\markboth{PUBLISHED IN IEEE TRANSACTIONS ON MAGNETICS. DOI: \href{https://doi.org/10.1109/TMAG.2024.3496695}{doi.org/10.1109/TMAG.2024.3496695}}%
{Chandra \MakeLowercase{\textit{et al.}}: Modeling Magnetic Hysteresis with Neural Operators}

\IEEEtitleabstractindextext{%
\begin{abstract}
Hysteresis modeling is crucial to comprehend the behavior of magnetic devices, facilitating optimal designs. Hitherto, deep learning-based methods employed to model hysteresis, face challenges in generalizing to novel input magnetic fields. This paper addresses the generalization challenge by proposing neural operators for modeling constitutive laws that exhibit magnetic hysteresis by learning a mapping between magnetic fields. In particular, three neural operators\textemdash deep operator network, Fourier neural operator, and wavelet neural operator\textemdash are employed to predict novel first-order reversal curves and minor loops, where novel means they are not used to train the model. In addition, a rate-independent Fourier neural operator is proposed to predict material responses at sampling rates different from those used during training to incorporate the rate-independent characteristics of magnetic hysteresis. The presented numerical experiments demonstrate that neural operators efficiently model magnetic hysteresis, outperforming the traditional neural recurrent methods on various metrics and generalizing to novel magnetic fields. The findings emphasize the advantages of using neural operators for modeling hysteresis under varying magnetic conditions, underscoring their importance in characterizing magnetic material based devices. The codes related to this paper are at \href{https://github.com/chandratue/magnetic_hysteresis_neural_operator}{github.com/chandratue/magnetic\_hysteresis\_neural\_operator}.

\end{abstract}

}

\maketitle
\IEEEdisplaynontitleabstractindextext
\IEEEpeerreviewmaketitle

\section{Introduction}
\IEEEPARstart{C}{ontemporary} industrialization thrives on magnetic material based devices. Those devices, such as electric machines and actuators, consume more than half of the global electrical energy. Optimizing the performance of these devices includes the estimation of energy dissipated in iron as losses, which is paramount for overall performance. A considerable component of iron loss calculation is the hysteresis loss, approximated through an accurate hysteresis model \cite{bertotti2005science, mayergoyz1986mathematical}. A material-specific hysteresis model precisely governs the constitutive relationship between the applied magnetic field ($H$) and the magnetic flux density ($B$). The relationship between these fields is nonlinear and history-dependent, traditionally modeled by the Jiles-Atherton \cite{jiles1986theory} and Preisach method \cite{preisach1935magnetische}, among others \cite{moree2023review}. Owing to the low inference cost and performance of machine learning-based methods in science and engineering, deep learning-based methods were introduced for modeling magnetic hysteresis in \cite{serpico1998magnetic} and are frequently applied in recent works, including but not limited to \cite{vuokila2023application, adly2021field, he2022improved}.

Among the deep learning based methods, recurrent neural architectures present a viable approach for modeling hysteresis due to their inherent reliance on historical data \cite{serrano2022neural, cao2006hysteresis, chandra2023neural, zhang2023magnetic, li2023magnet}, aligning with the history-dependent nature of hysteresis. However, these architectures primarily achieve accuracy for specific excitations, where predictions are limited to loops for which training has been performed. In real-world scenarios, however, where machine predictions are required for novel acting magnetic fields, generalization\textemdash which refers to the ability of deep learning models to predict outside of the training domain \cite{kapoor2024neural}\textemdash remains an open problem. Consequently, the limitation of current state-of-the-art neural architectures hinders the generalization across diverse excitation conditions and the estimation of responses for novel $H$ fields, making them unreliable as a part of the magnetic device model. 

The limitations of conventional neural networks, including recurrent neural networks (RNN) \cite{lecun2015deep} and their gated variants, such as long short-term memory (LSTM) \cite{hochreiter1997long} and gated recurrent unit (GRU) \cite{cho2014learning}, arise from their ability to learn only fixed-dimensional mappings between magnetic fields. These networks cannot model mappings between functions in continuous domains \cite{chen1995universal, azizzadenesheli2024neural}. This paper introduces neural operators \cite{azizzadenesheli2024neural, kovachki2023neural, lu2022comprehensive} to model the hysteresis relationship between magnetic fields to address these challenges. Unlike conventional neural networks that learn fixed-dimensional mappings, neural operators approximate the underlying operator, representing the mapping between $H$ and $B$ fields, to predict material responses ($B$ fields) for novel $H$ fields. Specifically, neural operators can approximate continuum mappings even when trained on discrete data, allowing them to generalize to novel $H$ fields.

This paper proposes to employ three neural operators\textemdash deep operator network (DeepONet) \cite{lu2021learning}, Fourier neural operator (FNO) \cite{li2020fourier}, and wavelet neural operator (WNO) \cite{tripura2023wavelet}\textemdash to model the $B$--$H$ hysteresis relationship, which has not been done before for the modeling of magnetic hysteresis. These operator networks are trained to predict novel first-order reversal curves (FORCs) and minor loops not included in the training data. The vanilla versions of \mbox{DeepONet}, FNO, and WNO are rate-dependent and consider the sampling rate during experiments. Further, to incorporate the rate-independent characteristic prevalent in magnetic hysteresis, this paper introduces a rate-independent Fourier neural operator (RIFNO) to predict $B$ fields when the sampling rates for the $H$ field differ between the training and test data. The primary contributions of this work are outlined as follows:

\begin{itemize}
\item This paper proposes an approach shift from learning a functional approximation between observed magnetic data to learning an operator between magnetic fields to achieve generalization, an important feature that neural magnetic hysteresis models must have.
\item The efficiency of neural operators is empirically demonstrated in the context of magnetic hysteresis modeling by comparing it with traditional recurrent neural architectures, highlighting their limitations. Specifically, the results are benchmarked against RNN, LSTM, GRU, and encoder-decoder LSTM (EDLSTM) models.
\item This paper proposes a rate-independent Fourier neural operator to incorporate rate-independent features of magnetic hysteresis and validate its performance through numerical experiments.
\item This paper bridges the gap between magnetic hysteresis modeling and scientific machine learning (SciML). We envision that the successful application of neural operators presented in this work will lead to employing and developing advanced neural operator techniques for magnetic hysteresis modeling.
\end{itemize}

The rest of the manuscript is organized as follows. Section~\ref{sec:motivation} presents the motivation to employ neural operators for modeling magnetic hysteresis. Section~\ref{sec:method} details the neural operator methodologies employed in this work. Section~\ref{sec:experiments} presents a series of numerical experiments conducted to demonstrate the accuracy of neural operators in modeling hysteresis. Section~\ref{sec:discussion} provides a discussion on the performance and prospects of the proposed approach. The main conclusions drawn from this study are collated in Section~\ref{sec:conclusions}.

\section{Motivation}\label{sec:motivation}
This section presents the motivation to advocate employing neural operators as surrogates for modeling the hysteresis phenomenon. Hitherto, neural network based function approximation strategies have been utilized to model hysteresis. More generally, traditional neural networks, such as feedforward or recurrent neural networks, leverage the universal function approximation property \cite{cybenko1989approximation, hornik1989multilayer, siegelmann1992computational}, ensuring that a continuous function \( f \) can be learned to approximate the given data pairs \((\bm{x}_i, y_i)\) such that \( y_i \approx f(\bm{x}_i) \) for all $N$ data points, \(1 \leq i \leq N\), \(i \in \mathbb{Z}\). Here, \( \bm{x}_i \) and \( y_i \) typically represent scalars, vectors, or tensors. The function \( f(\bm{x}_i) \) is learned as a surrogate for the underlying relationship via a neural network \( \mathcal{N}(\bm{x}_i, \theta) \), where \( \theta \) represents the network parameters optimized through a gradient-based optimization method to fit the data. The learned function \( f(\bm{x}_i) \) effectively maps the finite-dimensional data \( \bm{x}_i \) to \( y_i \), provided the optimizer finds a suitable local minima by minimizing an appropriate loss function.

In the context of scalar magnetic hysteresis, the data pairs \((\bm{x}_i, y_i)\) correspond to \((H_i, B_i)\), representing the applied magnetic field and induced magnetic field, respectively. These data pairs are often obtained through measurements or numerical models, such as the Preisach model, which acts as a surrogate for lab experiments \cite{vuokila2023application, khan2022generalizable}. The index \( i \) denotes the \(i^{\mathrm{th}}\) measurement data. The function \( f \) encapsulates the hysteresis relationship learned from the provided data pairs. This model is expected to effectively map the finite-dimensional inputs to finite-dimensional outputs when accurately fitted. However, for novel input fields represented by a distinct \( H_i \) sequence, the learned function \( f \) fails to approximate the underlying dynamics accurately, as is presented in this work, highlighting a limitation in current strategies.

\emph{Why does the function \( f \), approximated through a neural network, fails to predict \( B \) fields for novel \( H \) excitations?} The reason lies in the limited expressive power of traditional neural networks, which are designed to map tensors from the input space ($H$) to tensors in the output space ($B$) to learn functions but face challenges in generalizing this concept to map functions to functions \cite{lu2021learning}. The appropriate mathematical object that satisfies this objective is an operator. An operator \( \mathcal{D}: \mathcal{G} \rightarrow \mathcal{H} \) maps a function \( g \in \mathcal{G}\) to a function \( h \in \mathcal{H}\), where $\mathcal{G}$ and $\mathcal{H}$ are functional spaces. In the context of magnetic hysteresis, the functional spaces are the applied magnetic field ($H$) and the magnetic flux density ($B$). The hysteresis operator $\mathcal{D}\!: H \rightarrow B$ maps the excitation field functions $h \in  H$ to flux density functions $b \in B$. This mapping ensures that \( \forall h \in H, \quad \exists b \in B: \mathcal{D}(h) = b \).

Further, the fundamental difference between the operations of a vanilla neural network and an abstract neural operator could be described by considering the operations of a linear layer. A linear layer of a vanilla neural network maps \(\bm{x}\) to \(y\) through the operation, \(y = w \bm{x} + \beta\), where \(\bm{x}\) and \(y\) represent the set of data pairs \((\bm{x}_i, y_i)\). Precisely, \(\bm{x} \in \mathbb{R}^m\), \(y \in \mathbb{R}^n\), \(w \in \mathbb{R}^{n \times m}\), and \(\beta \in \mathbb{R}^n\). However, an abstract neural operator maps the function $h(\bm{x}) \in  H$ to a corresponding function $b(y) \in  B$ through the operation, \(b(y) = K(\bm{x}, y; w_k) h(\bm{x}) + \beta(\bm{x}; w_b)\), where \(w_k, w_b\) are the layer parameters and \(K(.)\) represents the transformation kernel which differs for each operator. For instance, for FNO and WNO, the kernel takes the form of Fourier and Wavelet integrals. In contrast, for DeepONet, the kernel corresponds to the dot product operation between the branch and trunk net, described in detail in the Methodology section.

Hence, neural operators could be understood as a generalization of vanilla neural networks to approximate a higher abstract object, known as operators, mapping functions to functions. In practice, it is challenging to model the hysteresis operator $\mathcal{D}$ analytically, motivating the development of neural surrogates for the operator, which are referred to as neural operators. 

Neural operators inherit the advantages of deep learning-based methods compared to the Preisach \cite{preisach1935magnetische} and the Jiles-Atherton model \cite{jiles1986theory}. One benefit of neural operators compared to the Preisach model is that it eliminates the need to rely on a specific set of measured data to characterize materials. In contrast, the classical Preisach model and its generalized versions necessitate the identification of particular branches \cite{serpico1998magnetic}. Another benefit of deep learning-based methods is their comparatively cheap inference cost compared to the Preisach model, as discussed in several works \cite{quondam2023deep, antonio2021effective, fulginei2012neural}. Neural operators in the inference step only require a forward pass of the architecture employing algebraic and transcendental operations compared to the integral operations in the Preisach model. Also, neural operators do not need to be degaussed \cite{boots2000anhysteretic, livshitz2011degaussing}, reducing the inference cost. The Jiles-Atherton model, a popular model for hysteresis modeling, faces challenges in approximating minor loops and non-symmetrical loops, as discussed in \cite{vijn2020parameter, atyia2024limitations}, among others. Also, the Jiles-Atherton model requires the physical parameters before characterizing the material, which is not required for neural operators.

The next section describes the methodologies for different neural operators employed in this paper.

\section{Methodology}\label{sec:method}
This section details the approach for modeling magnetic hysteresis through neural operators. As neural operators map functional spaces, the first step in modeling hysteresis involves the selection of input function space, which is the choice for excitation fields. The selection of input space dictates the functions utilized for training and testing the operator. The underlying assumption is that the trained neural operator can effectively generalize to any excitation field within the input functional space. This paper chooses two distinct input spaces corresponding to FORCs and minor loops. Additional details about the generation of functions in the input space are presented in Section~\ref{sec:experiments}.

Upon selecting the input space \( H \), excitation fields \(h\) are drawn from this space, \( h\! \sim H \). The space \( H \) can represent sinusoidal functions, Gaussian random fields, and others, depending on the specific application. The number of samples (\( N_{\text{sample}} \)) required to obtain a certain approximation accuracy, depends on the complexity of the functional space, the neural operator being employed, and the material under investigation. Since functions must be represented discretely on a computer or from data, the sampling is performed at \( t_{\text{sample}} \) locations. After generating \( N_{\text{sample}} \) excitation fields \( h\! \sim H\) at \( t_{\text{sample}} \) locations, these fields are applied to the material under study to produce the magnetic flux densities. This process yields \( N_{\text{sample}} \) instances of \( b \in B \) corresponding to each \( h\! \sim H \). Consequently, a raw dataset pair \((h_i, b_i)\) is formed, where $h_i$ and $b_i$ each represent a function for each $i$, $i = 1, \ldots, N_{\text{sample}}$. 

In cases where the raw dataset pair has to be generated by a Preisach-based model which takes as input $B$ fields and outputs the corresponding $H$ fields, like in this work, the dataset pair \((h_i, b_i)\) could be generated similarly by first sampling $B$ fields from an appropriate functional space and then generating the corresponding $H$ fields through the Preisach-based model.

The presented approach is intended to serve as a proof of concept demonstrating its effectiveness in learning and generalizing the hysteresis behavior using the Preisach model-based data \cite{daniels2023everett, 2410.02797}. Modeling materials based on data obtained from models draw similarities to standard practice in machine learning-based material modeling, where model-based data are used to develop neural surrogate models. For instance, in \cite{vuokila2023application, khan2022generalizable}, among others, authors utilize the Preisach model to generate data that are further used to train the neural network-based models. This work adopts this strategy to accelerate the development and validation of neural operators.

\begin{figure}[!ht]
  \centering
  \includegraphics[width=0.5\textwidth]{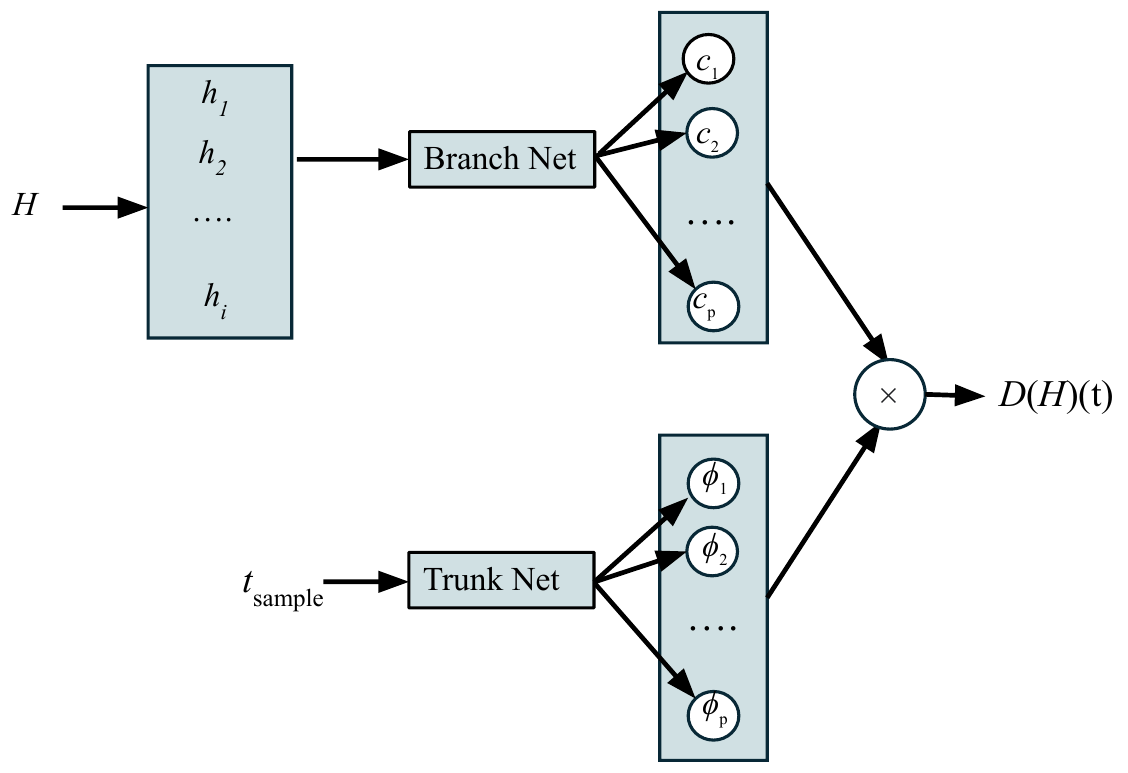}
  \caption{Deep operator network (DeepONet) architecture. The architecture consists of two separate feedforward neural networks\textemdash branch net and trunk net\textemdash whose outputs are combined using a dot product to approximate the $B$ fields.}
  \label{fig:deeponet_arch}
\end{figure}

Furthermore, neural operators have been employed for different science and engineering tasks except modeling magnetic hysteresis. The studies, including but not limited to \cite{colen2024interpreting, jafarzadeh2024peridynamic, kissas2022learning}, demonstrate that neural operators are effective even for experimental data. These advancements motivate the feasible future adoption of the proposed methodology in the magnetic domain for experimental data.

The raw dataset generated from the Preisach-based model \cite{daniels2023everett, 2410.02797} for this study is divided randomly into training and testing datasets. This results in two datasets \((h_i, b_i)\) and \((h_j, b_j)\), with \( 1 \leq i \leq N_{\text{train}} \), \( N_{\text{train}}+1 \leq j \leq N_{\text{train}} + N_{\text{test}} \). Here, \(N_{\text{train}} \) and \(N_{\text{test}} \) represent the number of training and test samples with the relation \(N_{\text{train}} + N_{\text{test}} = N_{\text{sample}} \). For the numerical experiments considered in this study, the number of training and test samples are kept equal, \(N_{\text{train}} = N_{\text{test}} \), to have a fair comparison with recurrent methods. Further details regarding the number of samples are presented in Section~\ref{sec:experiments}. The neural operators will be trained with the dataset \((h_i, b_i)\) and tested with \((h_j, b_j)\). In the rest of the paper, we omit the dependency on time, \textit{i.e.}, $h_i$ and $b_i$ can both denote the functions and those functions sampled at $t_\text{sample}$ locations. As the fields \(h_i\) and \(h_j\) are distinct, the operator methods provide a framework to generalize for novel input excitation fields.

The first neural operator employed in this work is the DeepONet. The standard DeepONet architecture \cite{lu2021learning} is built on fully connected feedforward neural networks as shown in Fig.~\ref{fig:deeponet_arch}. It comprises two distinct feedforward neural networks, the branch and trunk nets. In this study, the branch net takes functions from the input functional space: the training $H$ fields, $h_i$. The trunk net takes an array where the output fields are measured, specifically the $t_{\text{sample}}$ array. The branch and trunk nets include hidden layers and output neurons, similar to typical feedforward networks.

\begin{figure}[!ht]
  \centering
  \includegraphics[width=0.5\textwidth]{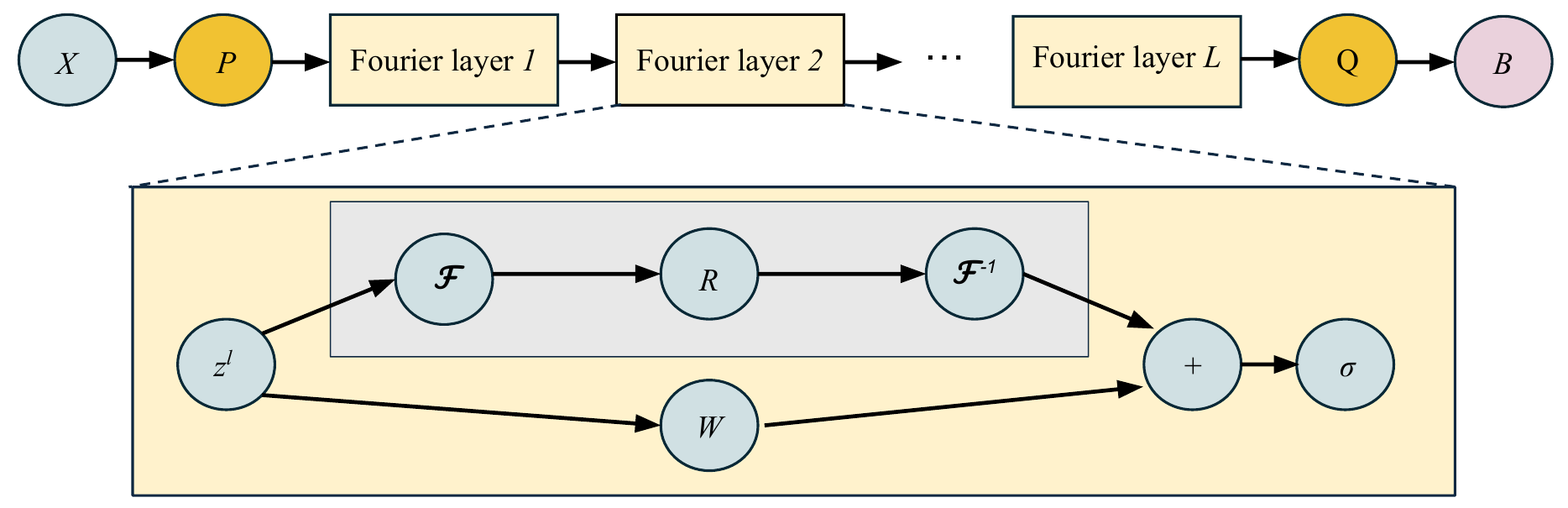}
  \caption{Neural architecture for Fourier neural operator (FNO) and rate-independent Fourier neural operator (RIFNO). For FNO the input $X\!:= [h_i, t_\text{sample}]$, whereas for RIFNO, \mbox{$X:=h_i$}. The input is passed through projection tensor ($P$) and Fourier layers and finally downscaled ($Q$) to approximate the $B$ field.}
  \label{fig:fourier_arch}
\end{figure}

While the number of hidden layers in the branch and trunk nets can vary, the number of outputs is identical. This requirement is necessary because both nets' outputs undergo a dot product operation to approximate the $b$ fields, $\hat{b}_i$. Mathematically, $b_i \approx \hat{b}_i = \sum_{k=1}^p c_k \phi_k$, where $\phi_k$ represents the outputs of the trunk net and $c_k$ represents the outputs of the branch net. From a conventional function approximation perspective, $\phi_k$ can be viewed as basis functions for approximating a function, and $c_k$ as the coefficients of these basis functions, combined over $p$ terms to approximate the field $b_i$. The mean-square loss function for DeepONet is,
\begin{equation}\label{eq1}
L(\theta) = \frac{1}{N_{\text{train}}} \sum_{i=1}^{N_{\text{train}}} (\hat{b}_i(\theta) - b_i)^2,
\end{equation}
where the network parameters $\theta$ are found through a gradient-based optimization method. Other loss functions minimizing the residual may also be considered using different norms \cite{kapoor2022predicting}. 

The second neural operator employed in this work is FNO, based on convolutional neural network architecture as shown in Fig.~\ref{fig:fourier_arch}. FNO does not employ separate networks to process the input fields ($h_i$) and $t_\text{sample}$ array; instead, it concatenates them by repeating the $t_\text{sample}$ array sufficiently to form the input tensor $X:= [h_i, t_\text{sample}]$ of channel size two. The input $X$ is then passed through a projection tensor, $P$,  to increase the number of channels to $N_\text{f}$, \textit{i.e.}, \(z^0 = P X\). The tensor $z^0$ then undergoes a sequence of Fourier layers, parameterized by learnable tensors $W$ and $R$. The computations for these Fourier layers can be expressed as,
\begin{equation} \label{eq2}
    z^{l+1} = \sigma (W^l z^l + \mathcal{F}^{-1} (R^l \mathcal{F} (z^l))),
\end{equation}
here $z^l$ denotes the latent tensor at $l^\text{th}$-layer. First, it undergoes a fast Fourier transform (FFT) $\mathcal{F}$. Subsequently, first $n_\text{m}$ modes are passed through a linear transformation  $R \in \mathbb{C}^{\mathcal{F}_{\text{out}} \times \mathcal{F}_{\text{out}}}$ followed by inverse FFT, $\mathcal{F}^{-1}$, where $\mathcal{F}_\text{out}$ refers to the dimension of $\mathcal{F}(z^l)$. The obtained tensor is summed with a linearly transformed latent tensor $W^l z^l$. Finally, a nonlinear activation function ($\sigma$) is applied to obtain $z^{l+1}$. These computations are repeated for $L$ number of Fourier blocks. The output of the last Fourier block $z^{L}$ is then downscaled by two projection tensors $Q$ and $\hat{Q}$, to obtain the predictions for the fields $b_i$,
\begin{equation}\label{eq3}
     \hat{b}_i = \hat{Q} \sigma (Q z^{L}).
\end{equation}

The same loss function~\eqref{eq1} as in the case of DeepONet is applied to approximate $b_i$.

\begin{figure}[!ht]
  \centering
  \includegraphics[width=0.5\textwidth]{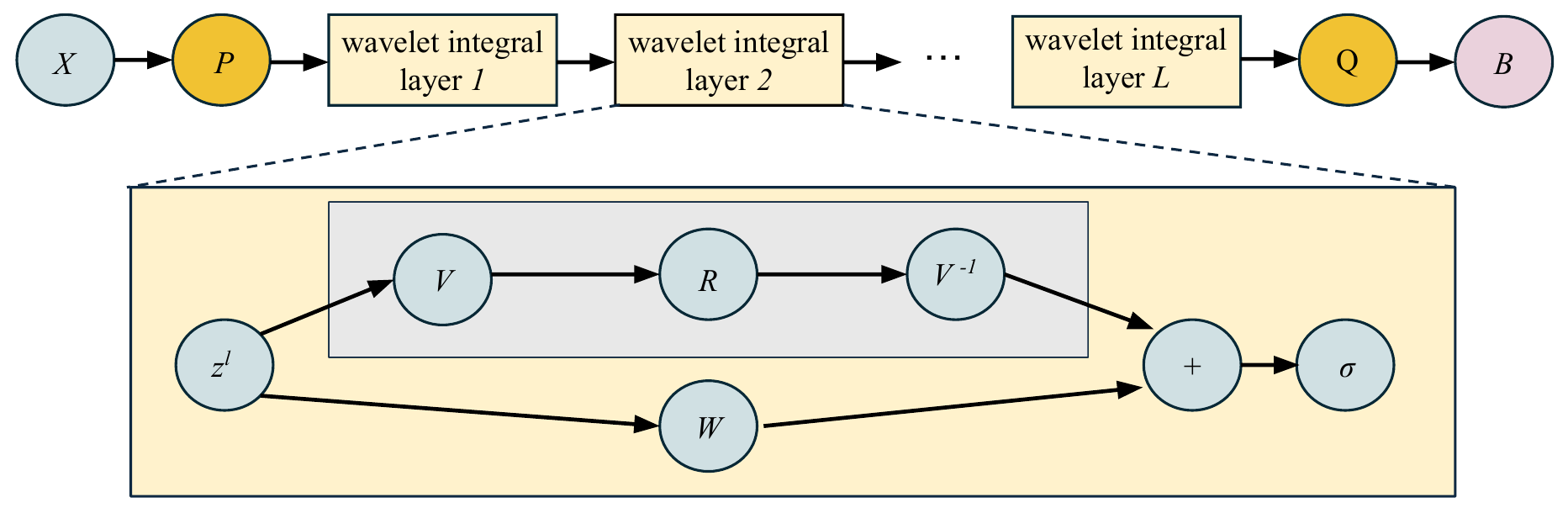}
  \caption{{Neural architecture for wavelet neural operator (WNO). The input $X\!:= [h_i, t_\text{sample}]$ is passed through projection tensor ($P$) and wavelet integral layers and finally downscaled ($Q$) to approximate the $B$ field.}}
  \label{fig:wno_arch}
\end{figure}

Similar to the universal function approximation property for the neural networks \cite{cybenko1989approximation, hornik1989multilayer, siegelmann1992computational}, DeepONet and FNO exhibit the universal approximation property for operators as presented in \cite{lu2021learning, chen1995universal, lanthaler2022error, kovachki2021universal}. This property ensures that DeepONet and FNO can approximate any continuous operator with arbitrary accuracy, provided sufficient network parameters. However, like the universal function approximation property for neural networks, the exact number of parameters and network configuration is unknown a priori.

In the context of this work, the magnetic hysteresis phenomenon, while complex, can be modeled as a continuous operator between the applied magnetic field and the material's magnetization response as described in Vol. I, Ch. 1 of \cite{bertotti2005science}. Since the approximations required for hysteresis behavior are continuous, the universal approximation theorem provides an established mathematical foundation for employing DeepONet and FNO to learn the mapping between magnetic fields, even under complex and varying field conditions, and ensuring accurate and reliable predictions for complex field variations.

The third neural operator employed in this work is the wavelet neural operator (WNO) \cite{tripura2023wavelet}, which leverages wavelet transforms to capture the hysteresis behavior. The architecture of WNO is similar to FNO, as shown in Fig.~\ref{fig:wno_arch}. However, WNO utilizes wavelet transforms instead of Fourier transforms to learn the underlying operator.

Similar to FNO, the input of WNO is the concatenation of the input field $h_i$ and $t_\text{sample}$. The concatenated tensor, $X := [h_i, t_\text{sample}]$, is passed through a projection tensor $P$ to increase the channel size and obtain $z^0$ similar to FNO. Subsequently, the latent tensor $z^0$ undergoes a series of wavelet integral layers, parameterized by learnable weight tensors $W$ and $R$. The operation of the wavelet layers can be written as:
\begin{equation} \label{eq:wno1}
 z^{l+1} = \sigma\left(W^l z^l + V^{-1}(R^l V(z^l))\right),
\end{equation}
where \( z^l \) represents the latent tensor at the $l^\text{th}$ layer, and $V$ denotes the wavelet transform, which decomposes the tensor into its wavelet coefficients. Instead of truncating certain modes (analogous to $n_\text{m}$ in FNO), WNO progressively decomposes the input signal into low-frequency approximations ($n_\textrm{l}$), to capture coarse and fine features of the signal. After applying the inverse wavelet transform, $V^{-1}$, the transformed tensor is summed with the linearly transformed latent tensor $W^l z^l$. A nonlinear activation function, $\sigma$, is applied to obtain $z^{l+1}$.

The transformations are repeated across $L$ wavelet integral layers. The output from the final layer, $z^L$, is passed through two projection tensors $Q$ and $\hat{Q}$ to reduce the dimensionality and produce the final predicted field $\hat{b}_i = \hat{Q} \sigma(Q z^L)$. The same loss function~\eqref{eq1}, used in the case of DeepONet and FNO, is applied to approximate the field outputs.

DeepONet, FNO, and WNO require the magnetic field's measurement grid, the $t_\text{sample}$ array. However, magnetic hysteresis often exhibits the property of being rate-independent \cite{xiao2024basic}. This paper proposes a minor modification in the FNO architecture to develop a rate-independent Fourier neural operator, namely RIFNO presented in Fig.~\ref{fig:fourier_arch}, to incorporate the rate-independent characteristic of magnetic hysteresis. The difference between FNO and RIFNO is the input tensor $X$, which in the case of RIFNO is defined as $X=h_i$. Specifically, RIFNO does not concatenate the $t_\text{sample}$ array with $h_i$; hence, the architecture remains invariant to the sampling rate, making it adequate for modeling rate-independent magnetic hysteresis. 

The neural operator RIFNO is founded on FNO, a universal approximator for continuous operators \cite{kovachki2021universal}. The universal approximation property is independent of the number of input channels in the architecture. Hence, even after removing the second channel of $t_\text{sample}$ array, the resulting RIFNO posits to approximate the rate-independent continuous magnetic hysteresis operator.

\section{Numerical Experiments} \label{sec:experiments}
This section presents the numerical experiments to validate the proposed approach in predicting novel FORCs and minor loops on which training has not been performed. The three neural operator methods are compared with traditional recurrent architectures RNN, LSTM, GRU, and recently proposed EDLSTM \cite{li2023magnet} to showcase the gain obtained with operators. The codes related to this paper are at \href{https://github.com/chandratue/magnetic_hysteresis_neural_operator}{github.com/chandratue/magnetic\_hysteresis\_neural\_operator}.

The data for the numerical experiments is generated using a Preisach-based method \cite{daniels2023everett, 2410.02797} modeling the material NO27-1450H. The measurement data for the Preisach-based method is obtained by a Brockhaus MPG 200 D soft-magnetic material tester, using an Epstein frame calibrated to correspond with the IEC standard. The data are obtained under quasi-DC excitation, indicating that the rate of change of the magnetic flux density is controlled such that any eddy current fields are negligible, and the static hysteresis behavior is obtained. The Epstein sample strips are obtained using spark eroding and cut in the rolling direction.

\subsection{Error metric}
Three error metrics are used to quantify the performance of the methods. First is the relative error in $\mathbb{L}_2$ norm, defined as,
\begin{equation} \label{eq4}
\mathcal{R} = \frac{\sqrt{\frac{1}{N_{\text{test}}} \sum_{j=1}^{N_{\text{test}}} (\hat{b}_j - b_j)^2}}{\sqrt{\frac{1}{N_{\text{test}}} \sum_{j=1}^{N_{\text{test}}} b_j^2}}.
\end{equation}

The second error metric is the mean absolute error (MAE), used to quantify the average absolute difference between the predicted field $\hat{b}_j$ and the actual field $b_j$, defined as,
\begin{equation} \label{eq5}
\text{MAE} = \frac{1}{N_{\text{test}}} \sum_{j=1}^{N_{\text{test}}} |\hat{b}_j - b_j|.
\end{equation}

The third error metric is the root mean squared error (RMSE) used to quantify the average magnitude of the errors between predicted values $\hat{b}_j$ and actual values $b_j$, defined as,
\begin{equation} \label{eq6}
\text{RMSE} = \sqrt{\frac{1}{N_{\text{test}}} \sum_{j=1}^{N_{\text{test}}} (\hat{b}_j - b_j)^2}.
\end{equation}

\begin{figure*}[!t]
    \centering
    \begin{subfigure}[b]{0.15\textwidth}
        \includegraphics[width=\textwidth]{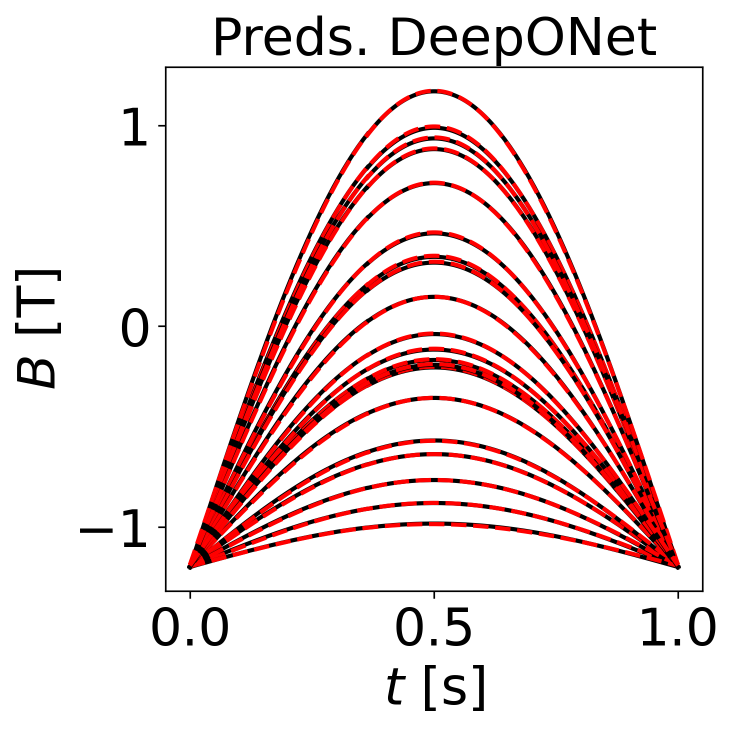}
    \end{subfigure}
    \begin{subfigure}[b]{0.15\textwidth}
        \includegraphics[width=\textwidth]{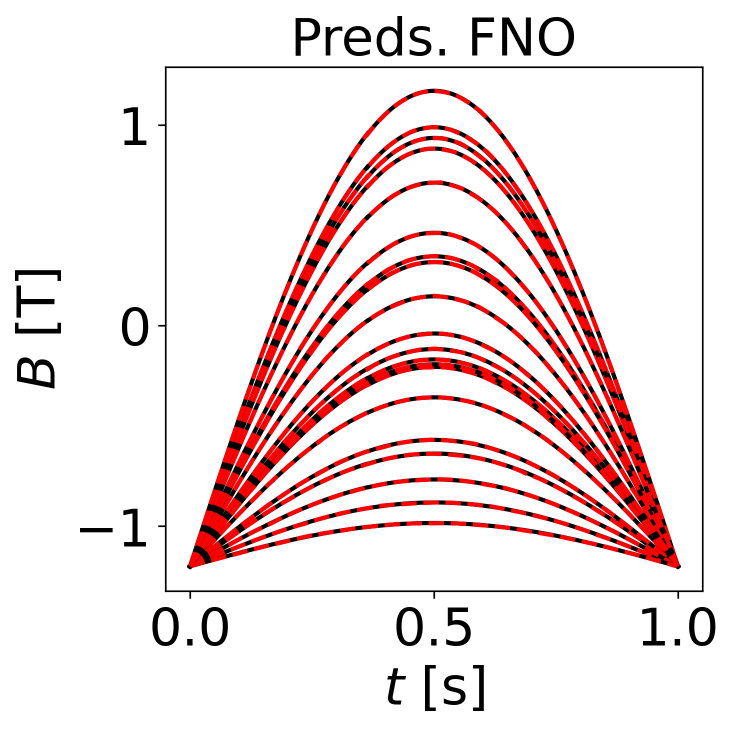}
    \end{subfigure}
    \begin{subfigure}[b]{0.15\textwidth}
        \includegraphics[width=\textwidth]{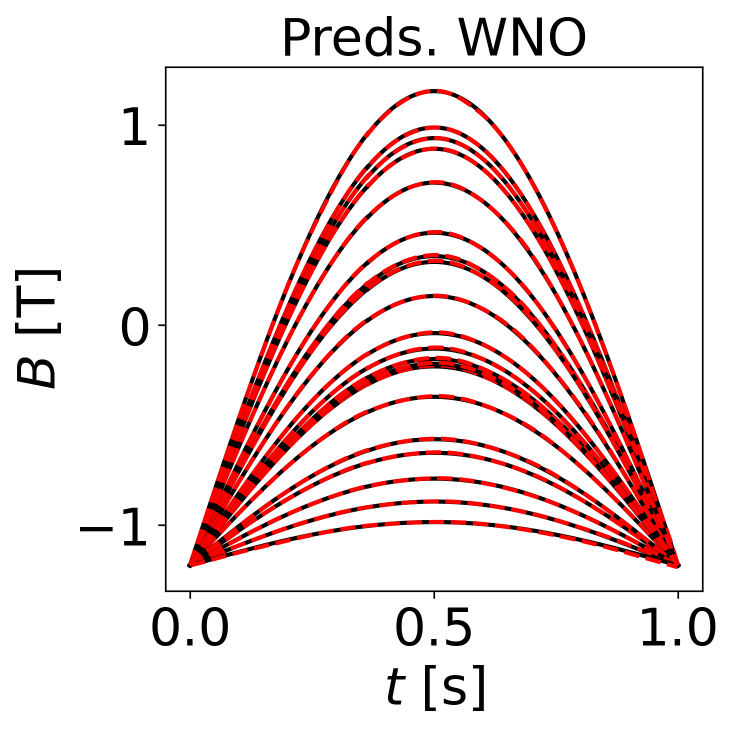}
    \end{subfigure}
    \begin{subfigure}[b]{0.15\textwidth}
        \includegraphics[width=\textwidth]{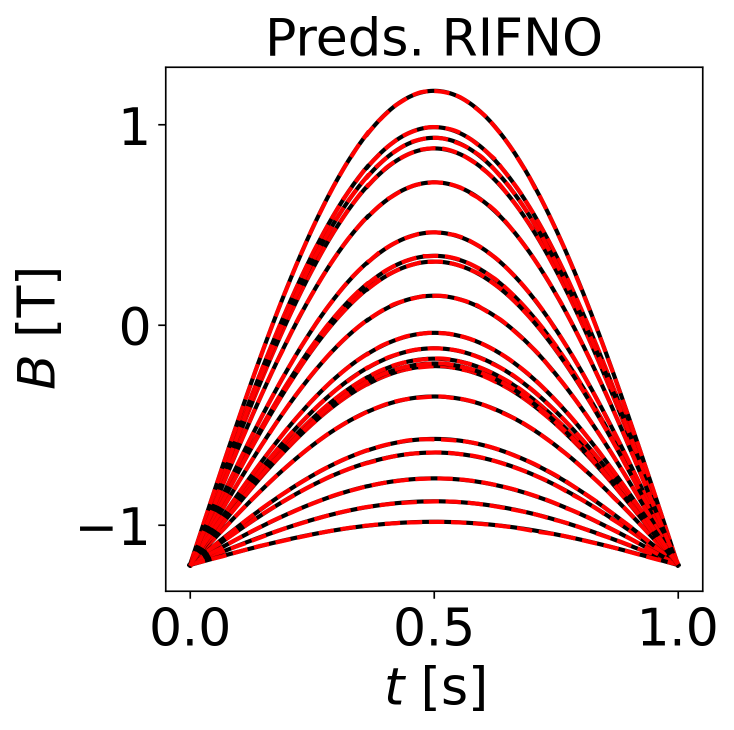}
    \end{subfigure}
    \begin{subfigure}[b]{0.15\textwidth}
        \includegraphics[width=\textwidth]{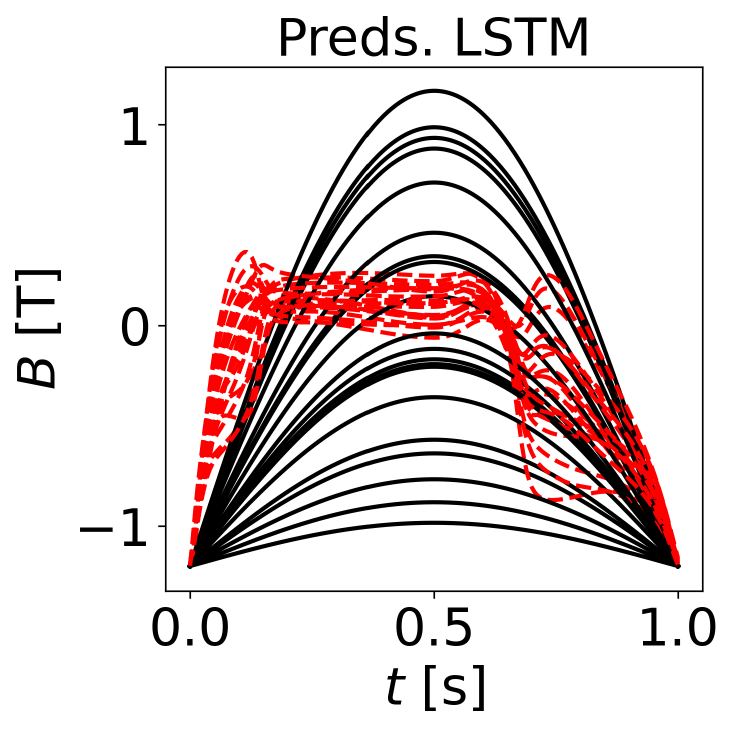}
    \end{subfigure}
    \begin{subfigure}[b]{0.15\textwidth}
        \includegraphics[width=\textwidth]{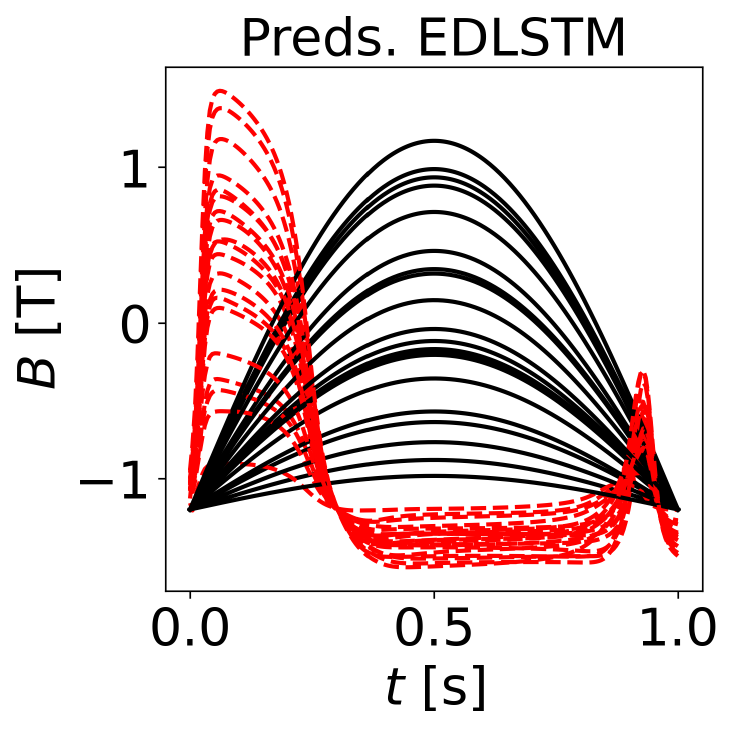}
    \end{subfigure}

    \vspace{0.25cm}

    \begin{subfigure}[b]{0.15\textwidth}
        \includegraphics[width=\textwidth]{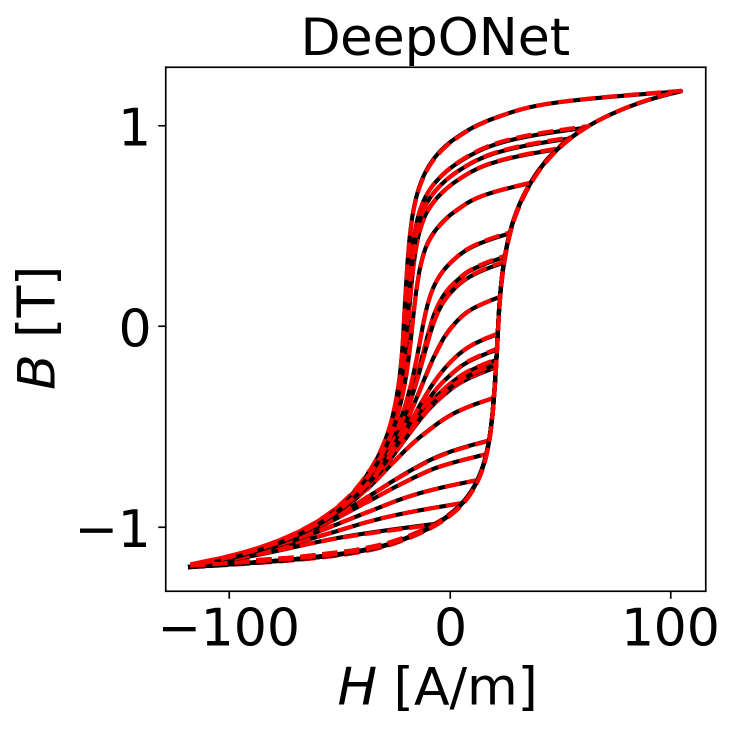}
    \end{subfigure}
    \begin{subfigure}[b]{0.15\textwidth}
        \includegraphics[width=\textwidth]{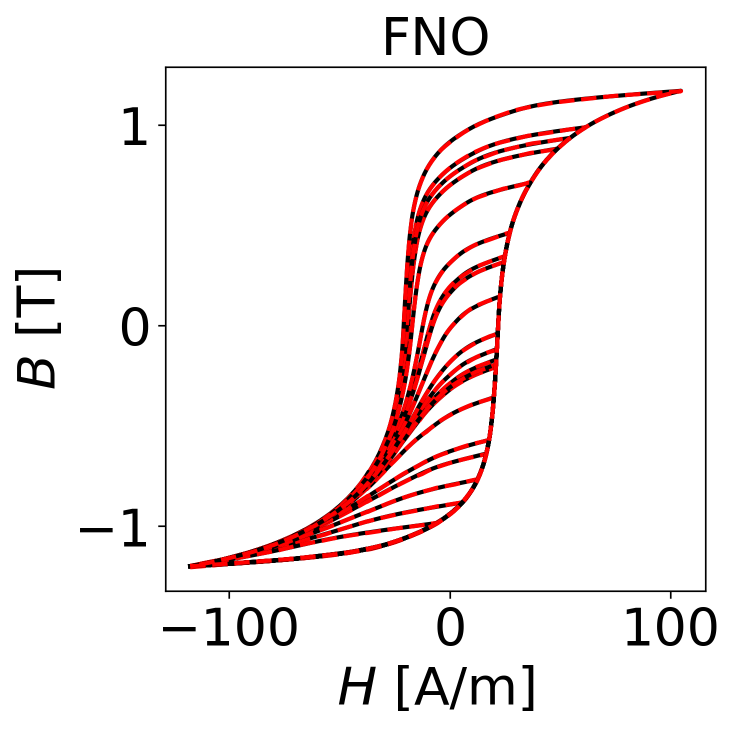}
    \end{subfigure}
    \begin{subfigure}[b]{0.15\textwidth}
        \includegraphics[width=\textwidth]{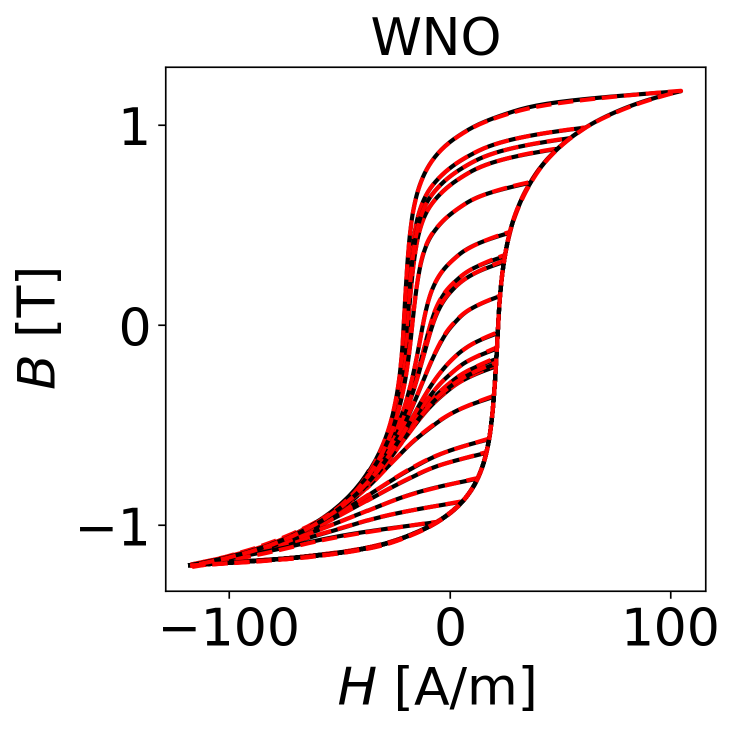}
    \end{subfigure}
    \begin{subfigure}[b]{0.15\textwidth}
        \includegraphics[width=\textwidth]{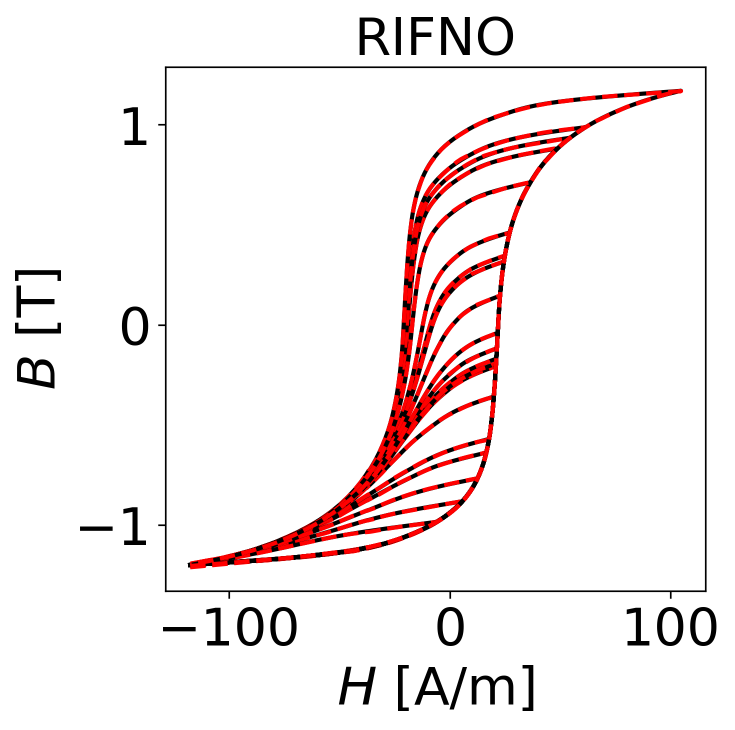}
    \end{subfigure}
    \begin{subfigure}[b]{0.15\textwidth}
        \includegraphics[width=\textwidth]{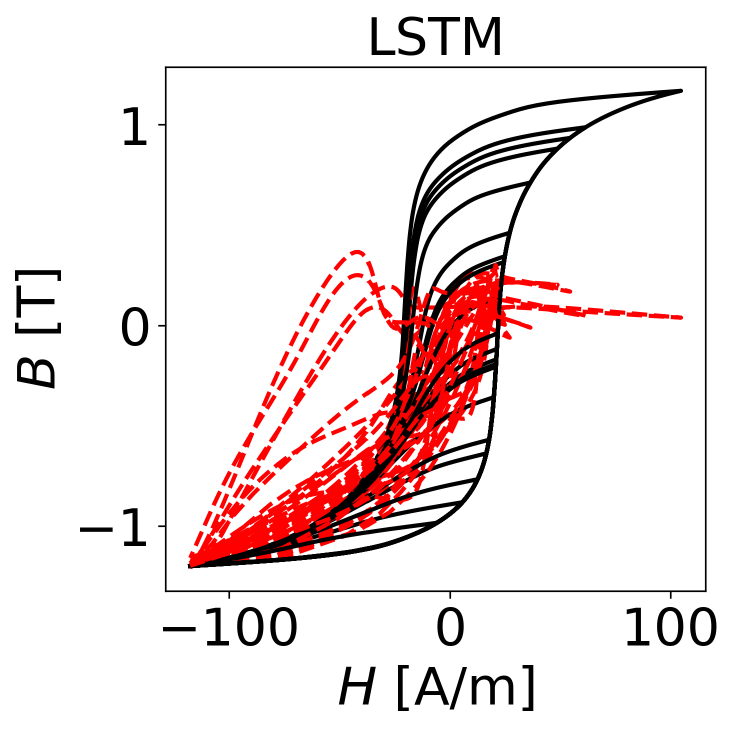}
    \end{subfigure}
    \begin{subfigure}[b]{0.15\textwidth}
        \includegraphics[width=\textwidth]{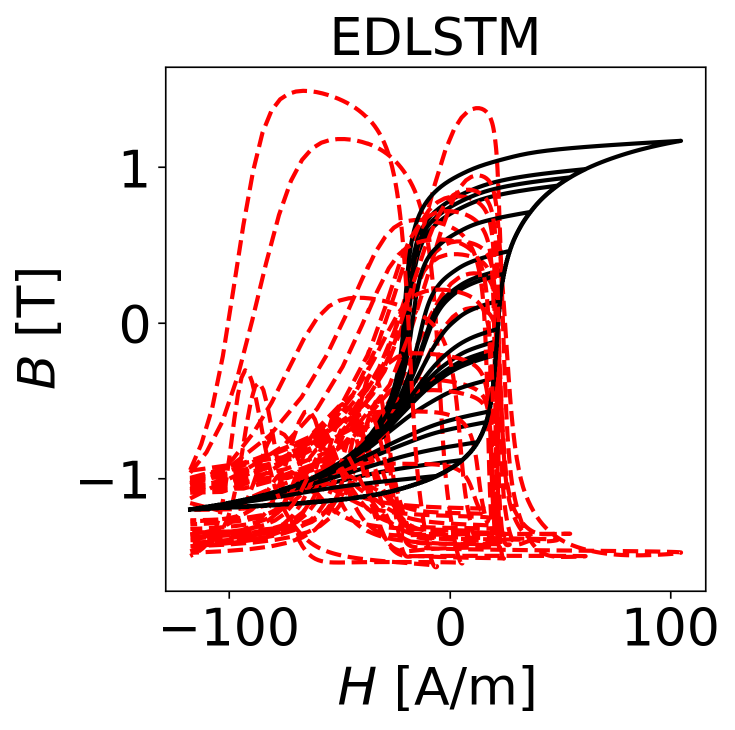}
    \end{subfigure}

    \vspace{0.25cm}

    \begin{subfigure}[b]{0.15\textwidth}
        \includegraphics[width=\textwidth]{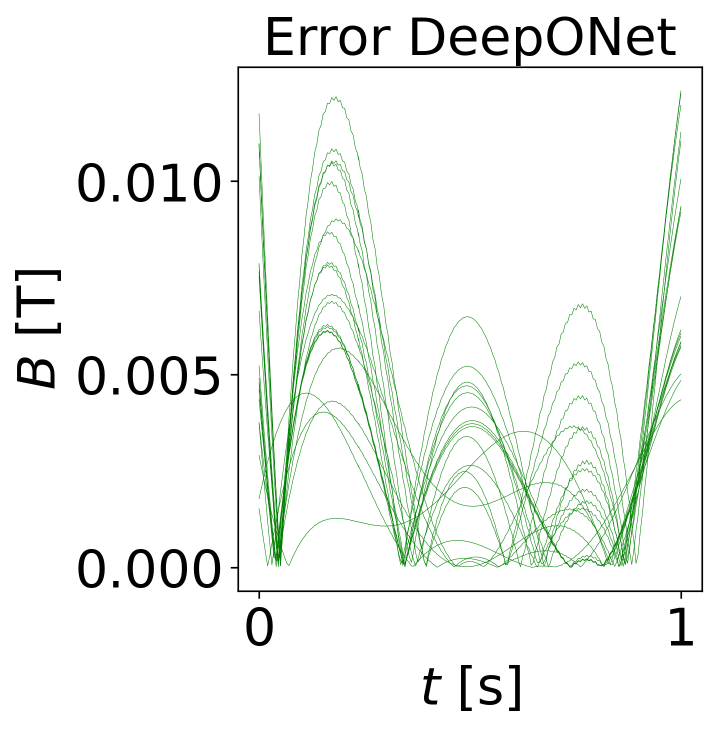}
    \end{subfigure}
    \begin{subfigure}[b]{0.15\textwidth}
        \includegraphics[width=\textwidth]{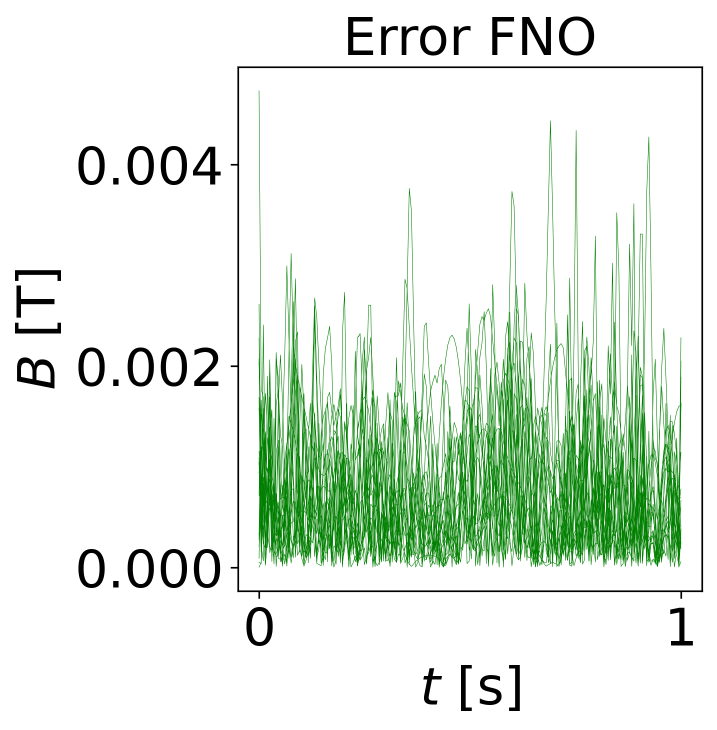}
    \end{subfigure}
    \begin{subfigure}[b]{0.15\textwidth}
        \includegraphics[width=\textwidth]{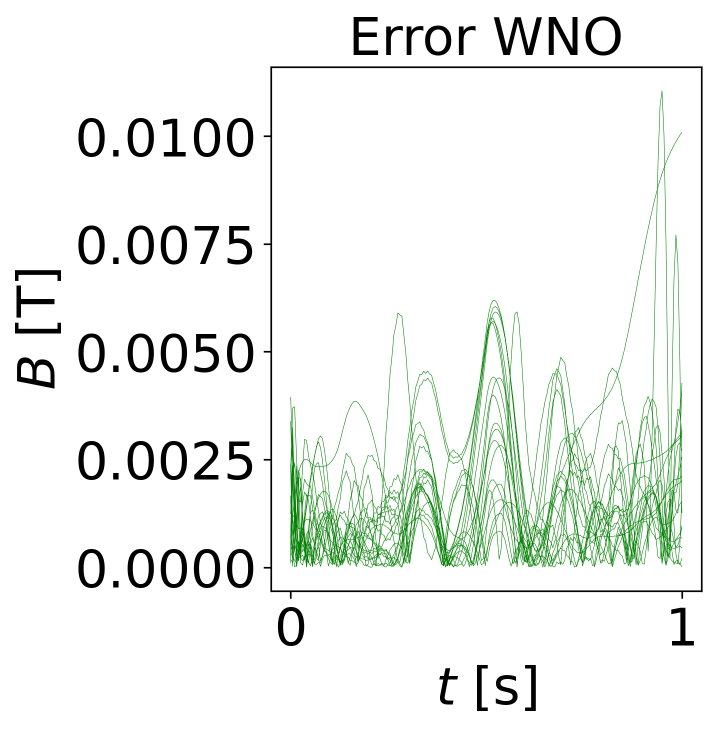}
    \end{subfigure}
    \begin{subfigure}[b]{0.15\textwidth}
        \includegraphics[width=\textwidth]{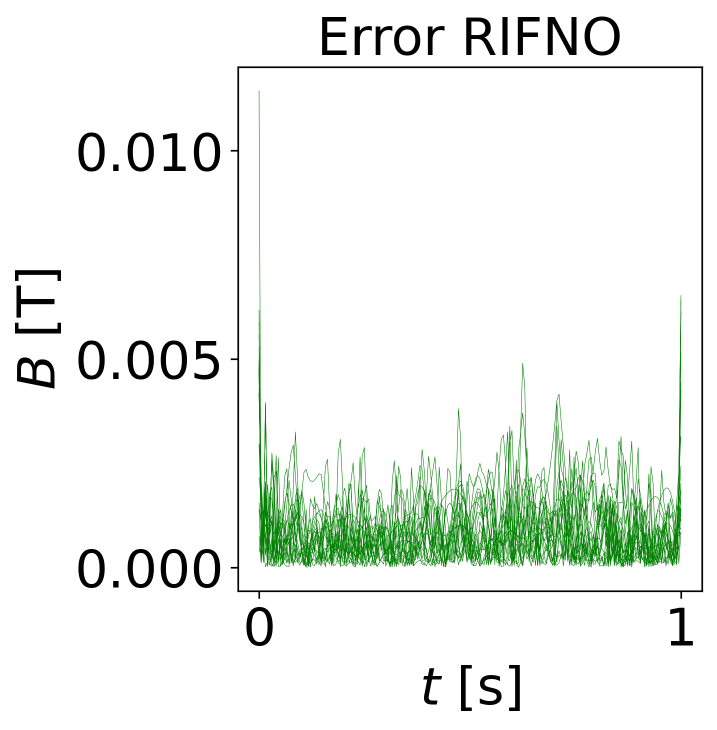}
    \end{subfigure}
    \begin{subfigure}[b]{0.15\textwidth}
        \includegraphics[width=\textwidth]{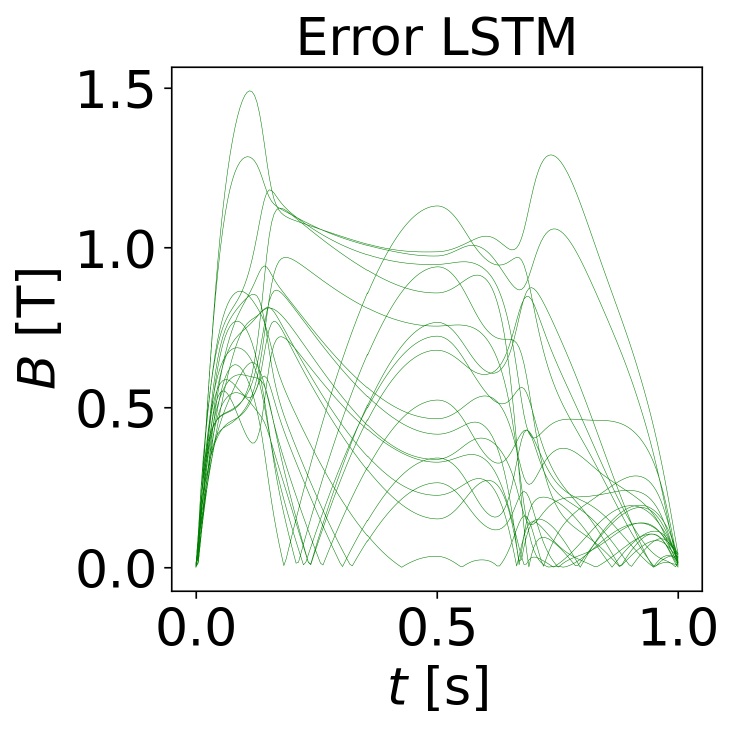}
    \end{subfigure}
    \begin{subfigure}[b]{0.15\textwidth}
        \includegraphics[width=\textwidth]{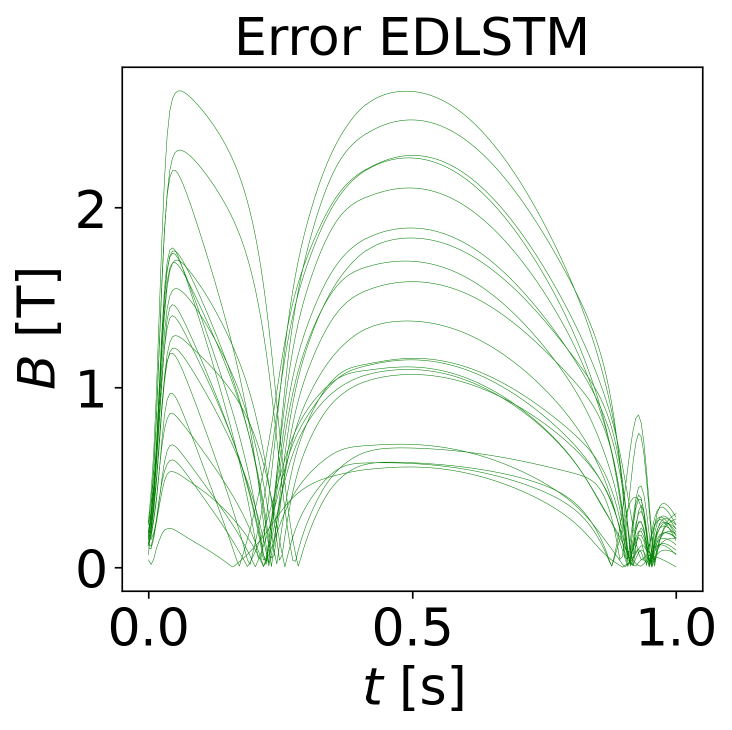}
    \end{subfigure}
    \caption{Performance of neural networks on predicting the first-order reversal curves. \textbf{Top row:} (From Left) the predictions (preds.) of the $B$ fields by DeepONet, FNO, WNO, RIFNO, LSTM, and Encoder-Decoder LSTM (EDLSTM) respectively. \textbf{Middle row:} (From Left) the predicted hysteresis loops (in red) compared with the reference (ref.) loops (in black) for the six methods. \textbf{Bottom row:} (From Left) absolute errors in predicting the FORCs for the six methods (in green).}
    \label{fig:FORC}
\end{figure*}

\subsection{Hyperparameter selection}
A hyperparameter search is conducted to predict the FORCs, and the obtained optimal hyperparameters are used consistently across all other cases.

For DeepONet, two key hyperparameters influencing the accuracy of the learned operator are the number of layers in the branch and trunk networks and the number of terms in the operator approximation \(p\). A popular hyperparameter selection strategy, grid search \cite{mishra2023estimates} is performed, varying the number of layers between 2, 4, 6, and 8, and the values of \(p\) between 25, 50, 75, and 100, resulting in 16 combinations. As shown in Table~\ref{tab:hyp_opt_don}, the optimal configuration, yielding the lowest error, corresponds to 8 layers and \(p = 25\).

For FNO, the key hyperparameters are the number of Fourier modes to retain (\(n_\mathrm{m}\)) and the number of channels for the projection \(N_\mathrm{f}\). Similar to DeepONet, a grid search is conducted by varying \(n_\mathrm{m}\) between 1, 2, 3, and 4, and \(N_\mathrm{f}\) between 2, 4, 6, and 8, leading to 16 total combinations. Table~\ref{tab:hyp_opt_fno} presents the results, with the best configuration identified as four modes and eight channels.

Similarly, for WNO, the important hyperparameters include the level of wavelet decomposition (\(n_\mathrm{l}\)) and the number of wavelet integral blocks \(L\). A grid search is conducted by varying \(n_\mathrm{l}\) between 1, 2, 3, and 4, and \(L\) between 2, 4, 6, and 8, resulting in 16 cases. The optimal configuration, as shown in Table~\ref{tab:hyp_opt_wno}, corresponded to four levels of decomposition and eight wavelet integral blocks.

The hyperparameter searches ensure the robustness of the presented results and provide optimal hyperparameter selection among the chosen cases for each neural operator.

\begin{table}[ht]
    \centering
    \caption{Hyperparameter selection for DeepONet}
    \begin{tabular}{|l|l|c|c|c|c|c|}
        \hline
        \multicolumn{2}{|c|}{} & \multicolumn{4}{c|}{Layers} \\ \hline 
        $p$ & Metric & 2 & 4 & 6 & 8 \\ \hline
     
        \multirow{3}{*}{25} & $\mathcal{R}$ & 4.96e-2 & 3.11e-3 & 1.91e-2 & \textbf{2.49e-3} \\ \cline{2-6}
                           & MAE  & 3.19e-2 & 1.72e-3 & 1.10e-2 & \textbf{1.49e-3} \\ \cline{2-6}
                           & RMSE & 3.58e-2 & 2.25e-3 & 1.38e-2 & \textbf{1.79e-3} \\ \hline
        \multirow{3}{*}{50} & $\mathcal{R}$ & 9.49e-3 & 4.4e-3 & 3.32e-3 & 4.34e-3 \\ \cline{2-6}
                           & MAE  & 4.96e-3 & 2.44e-3 & 1.89e-3 & 2.34e-3 \\ \cline{2-6}
                           & RMSE & 6.86e-3 & 3.17e-3 & 2.40e-3 & 3.13e-3 \\ \hline
        \multirow{3}{*}{75} & $\mathcal{R}$ & 2.10e-2 & 5.41e-3 & 4.93e-3 & 3.00e-2 \\ \cline{2-6}
                           & MAE  & 1.13e-2 & 3.00e-3 & 2.71e-3 & 2.07e-2 \\ \cline{2-6}
                           & RMSE & 1.51e-2 & 3.90e-3 & 3.56e-3 & 2.17e-2 \\ \hline
        \multirow{3}{*}{100} & $\mathcal{R}$ & 1.73e-2 & 8.91e-2 & 1.4e-2 & 5.40e-3 \\ \cline{2-6}
                            & MAE  & 9.49e-3 & 6.01e-2 & 9.91e-3 & 2.86e-3 \\ \cline{2-6}
                            & RMSE & 1.25e-2 & 6.44e-2 & 1.05e-2 & 3.90e-3 \\ \hline
    \end{tabular}
    \label{tab:hyp_opt_don}
\end{table}

\begin{table}[ht]
    \centering
    \caption{Hyperparameter selection for FNO}
    \begin{tabular}{|l|l|c|c|c|c|c|}
        \hline
         \multicolumn{2}{|c|}{} & \multicolumn{4}{c|}{Fourier modes ($n_\textrm{m}$)} \\ \hline
        $N_\textrm{f}$ & Metric & 1 & 2 & 3 & 4 \\ \hline
     
        \multirow{3}{*}{2} & $\mathcal{R}$ & 1.51e-1 & 5.47e-3 & 8.31e-1 & 1.16e-2 \\ \cline{2-6}
                           & MAE  & 5.37e-2 & 2.95e-3 & 5.07e-1 & 6.49e-3 \\ \cline{2-6}
                           & RMSE & 1.09e-1 & 3.95e-3 & 6.00e-1 & 8.44e-3 \\ \hline
        \multirow{3}{*}{4} & $\mathcal{R}$ & 1.69e-2 & 4.08e-3 & 3.02e-3 & 3.84e-3 \\ \cline{2-6}
                           & MAE  & 8.92e-3 & 2.27e-3 & 1.70e-3 & 2.12e-3 \\ \cline{2-6}
                           & RMSE & 1.22e-2 & 2.95e-3 & 2.18e-3 & 2.78e-3 \\ \hline
        \multirow{3}{*}{6} & $\mathcal{R}$ & 1.11e-2 & 2.40e-3 & 1.96e-3 & 1.95e-3 \\ \cline{2-6}
                           & MAE  & 6.30e-3 & 1.34e-3 & 1.13e-3 & 1.12e-3 \\ \cline{2-6}
                           & RMSE & 8.20e-3 & 1.73e-3 & 1.41e-3 & 1.41e-3 \\ \hline
        \multirow{3}{*}{8} & $\mathcal{R}$ & 8.31e-3 & 2.25e-3 & 1.73e-3 & \textbf{1.34e-3} \\ \cline{2-6}
                            & MAE  & 4.47e-3 & 1.28e-3 & 9.88e-4 & \textbf{7.84e-4} \\ \cline{2-6}
                            & RMSE & 6.00e-3 & 1.63e-3 & 1.25e-3 & \textbf{9.74e-4} \\ \hline
    \end{tabular}
    \label{tab:hyp_opt_fno}
\end{table}

\begin{table}[ht]
    \centering
    \caption{Hyperparameter selection for WNO}
    \begin{tabular}{|l|l|c|c|c|c|c|}
        \hline
         \multicolumn{2}{|c|}{} & \multicolumn{4}{c|}{Level of wavelet decomposition ($n_\mathrm{l}$)} \\ \hline
        $L$ & Metric & 1 & 2 & 3 & 4 \\ \hline
     
        \multirow{3}{*}{2} & $\mathcal{R}$ & 1.12e-1 & 2.38e-2 & 2.10e-2 & 1.59e-2 \\ \cline{2-6}
                           & MAE  & 3.21e-2 & 1.12e-2 & 9.61e-3 & 8.00e-3 \\ \cline{2-6}
                           & RMSE & 8.15e-2 & 1.72e-2 & 1.51e-2 & 1.15e-2 \\ \hline
        \multirow{3}{*}{4} & $\mathcal{R}$ & 3.18e-2 & 6.06e-3 & 1.88e-2 & 4.67e-3   \\ \cline{2-6}
                           & MAE  & 1.62e-2 & 3.13e-3 & 9.91e-3 & 2.46e-3\\ \cline{2-6}
                           & RMSE & 2.29e-2 & 4.38e-3 & 1.36e-2 & 3.37e-3 \\ \hline
        \multirow{3}{*}{6} & $\mathcal{R}$ & 1.06e-2 & 5.15e-3 & 4.01e-2 & 7.60e-3 \\ \cline{2-6}
                           & MAE  & 5.73e-3 & 2.83e-3 & 2.22e-2 & 4.59e-3 \\ \cline{2-6}
                           & RMSE & 7.66e-3 & 3.72e-2 & 2.90e-2 & 5.49e-3 \\ \hline
        \multirow{3}{*}{8} & $\mathcal{R}$ & 1.48e-2 & 2.10e-2 & 2.15e-2 &  \textbf{2.56e-3}\\ \cline{2-6}
                            & MAE  & 7.79e-3 & 1.28e-2 & 1.39e-2 & \textbf{1.41e-3}  \\ \cline{2-6}
                            & RMSE & 1.07e-2 & 1.52e-2 & 1.55e-2 & \textbf{1.85e-3} \\ \hline
    \end{tabular}
    \label{tab:hyp_opt_wno}
\end{table}

\subsection{First-order reversal curves}
The first numerical experiment entails predicting the FORCs. This study employs the Preisach model to generate the dataset for training and testing the neural networks \cite{daniels2023everett, 2410.02797}. The model inputs the magnetic flux density ($B$ field) and produces the corresponding magnetic field strength ($H$ field). The $b$ fields are sampled from a functional space $B$, used as the input for the Preisach model to produce the corresponding $H$ fields. This process generates data pairs ($h_i, b_i$), which are used to train the neural networks. Whether one uses experimental data or any other surrogate model characterizing the material, this approach remains invariant to the choice of input sample space and output generated fields.

To generate the dataset for training and testing the neural networks, the functional space of $B$ comprises half sine curves with amplitudes randomly sampled between \SI{0.1}{\tesla} and \SI{1.2}{\tesla}, and defined over the interval $[0, 1]$ with $t_{\text{sample}} = 198$. A total of 2000 sine curves are generated and distributed independently and identically between the training and testing sets, such that $N_{\text{train}} + N_{\text{test}} = 1000 + 1000 = 2000 = N_{\text{sample}}$. Before inputting the data into the neural networks, they are normalized to the range of -1 to 1 using min-max scaling, as described in \cite{chandra2023neural}, to facilitate the training process \cite{kapoor2023physics}.

The first neural operator utilized for modeling FORCs is the DeepONet. The operator uses 1000 random functions as input to the branch net and the $t_{\text{sample}}$ array of size 198 as input to the trunk net. The branch and trunk nets are fully connected feedforward neural networks, each comprising eight hidden layers with 200 neurons in each layer and a tanh activation function. The output dimension of both the branch and trunk nets is $p=25$, following the hyperparameter selection experiment. The model parameters are initialized using Xavier initialization \cite{glorot2010understanding}, and the network is trained using the ADAM optimizer with a learning rate of $5 \times 10^{-5}$ over 10000 epochs. 

\begin{table}[ht]
    \centering
    \caption{Errors in predicting first-order reversal curves ($\mathcal{R}$: relative error in $\mathbb{L}_2$ norm; MAE: mean absolute error; RMSE: root mean squared error.)}
    \begin{tabular}{|l|c|c|c|}
        \hline
        Method   & $\mathcal{R}$ & MAE & RMSE \\ \hline
        DeepONet & 2.49e-3       & 1.49e-3  & 1.79e-3  \\ \hline
        FNO      & \textbf{1.34e-3} & \textbf{7.48e-4}  & \textbf{9.74e-4}  \\ \hline
        WNO      & 2.56e-3       & 1.41e-3  & 1.85e-3  \\ \hline
        RIFNO    & 1.63e-3       & 8.79e-4  & 1.17e-3   \\ \hline
        RNN      & 1.19e+0       & 6.99e-1  & 8.61e-1  \\ \hline
        LSTM     & 8.08e-1       & 4.76e-1  & 5.83e-1  \\ \hline
        GRU      & 9.27e-1       & 5.18e-1  & 6.70e-1   \\ \hline
        EDLSTM   &1.64e+0        &9.81e-1   & 1.18e+0   \\ \hline
    \end{tabular}
    \label{tab:forc}
\end{table}

Further, FNO and RIFNO are employed to predict FORCs. Both architectures share identical structures with variations in input configuration. Specifically, FNO inputs concatenated data consisting of 1000 instances of $H$ fields and a sampling array repeated across 1000 samples, while RIFNO solely utilizes the 1000 instances of $H$ fields as input. Common hyperparameters across both networks include an initial fully connected layer projecting inputs from 2 channels to 8 channels, \textit{i.e.}, $N_\text{f} = 8$ , followed by four spectral convolutional layers ($L=4$), each with eight input and eight output channels. Within each block, the first four modes after Fourier transformation are retained ($n_m = 4$). The networks conclude with two fully connected layers transforming dimensions from 8 channels to 128 and finally to a single output channel. ReLU activation function is applied throughout the network. Training parameters encompass a batch size of 100 samples over 10000 epochs, utilizing the ADAM optimizer with a learning rate of $1 \times 10^{-4}$.

Furthermore, WNO is employed to predict FORCs. WNO inputs are identical to FNO, and are concatenated data consisting of 1000 instances of $H$ fields and a sampling array repeated across 1000 samples. An initial fully connected layer is utilized to project inputs from 2 channels to 64 channels, \textit{i.e.}, $N_\text{f} = 64$ , followed by eight wavelet integral layers ($L=8$), each with 64 input and 64 output channels. Within each block, four levels of wavelet decomposition ($n_\text{l} = 4$) are performed employing db6 wavelet. The networks conclude with two fully connected layers transforming dimensions from 64 channels to 128 and finally to a single output channel. GeLU activation function is applied throughout the network. Training parameters encompass a batch size of 100 samples over 10000 epochs, utilizing the ADAM optimizer with a learning rate of $1 \times 10^{-3}$.

The neural operators are subsequently benchmarked against traditional recurrent architectures, specifically RNN, LSTM, GRU, and recently proposed encoder-decoder LSTM (EDLSTM) \cite{li2023magnet}. Following the data processing similar to \cite{li2023magnet, chandra2023neural}, the input and output dimensions are set to 1000 for each cell. The sequence length is fixed at 198, corresponding to the size of the sampling array, $t_\text{sample}$. Each recurrent cell consists of 128 neurons with a single hidden layer. Training spans 10000 epochs using the ADAM optimizer with a learning rate \mbox{$1 \times 10^{-4}$}.

\begin{figure}[!ht]
  \centering
  \includegraphics[width=0.45\textwidth]{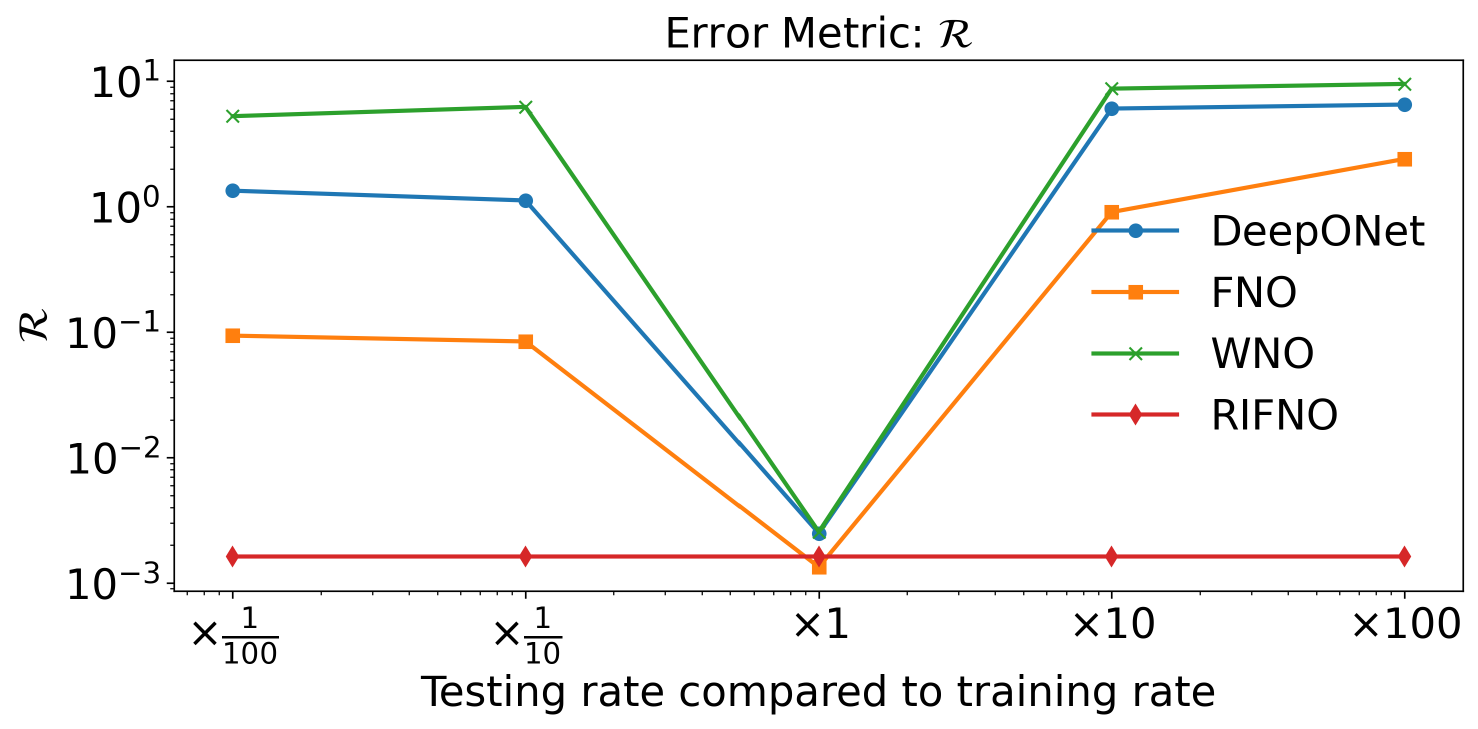}
  \caption{Variation of relative error with respect to different testing rates for neural operators.}
  \label{fig:line_curve}
\end{figure}

Furthermore, neural operators are compared with EDLSTM, a recurrent neural architecture to model magnetic hysteresis, as presented in \cite{li2023magnet}. The architecture comprises an encoder and a decoder. The encoder consists of a 1-layer 128-state LSTM network, processing the input sequence by leveraging the LSTM gate mechanisms, allowing the encoder to capture and store the characteristics of the input sequence as hidden and cell states. The decoder mirrors the architecture of the encoder. It generates the output sequence through an autoregressive inference method, iteratively updating by feeding the previous output back into the network to predict the next time step until the entire output sequence is generated. Training is carried out for 10000 epochs using the ADAM optimizer with a learning rate \mbox{$1 \times 10^{-4}$} as in the case of LSTM.

Fig.~\ref{fig:FORC} illustrates the results for the neural operators and sequential model in predicting magnetic hysteresis. The first row shows the predictions from six models: DeepONet, FNO, WNO, RIFNO, LSTM, and EDLSTM, with FNO demonstrating the best approximation. The second row compares the FORC predictions for each model, where red curves indicate predictions and black curves show the ground truth. The third row presents absolute error plots, showcasing that LSTM, and EDLSTM perform poorly with widespread errors. The error for DeepONet propogates sinusoidally, FNO and WNO have a maximum error around 0.004 and 0.01, respectively, and the error in RIFNO predictions are prominent near the ends. Overall, neural operators perform better in approximating magnetic hysteresis than the recurrent neural methods.

Additionally, Table~\ref{tab:forc} compares the prediction errors for FORCs across eight models: DeepONet, FNO, WNO, RIFNO, RNN, LSTM, GRU, and EDLSTM using three metrics: relative error (\(\mathcal{R}\)), MAE, and RMSE. FNO exhibits the lowest errors (\(\mathcal{R}\) = 1.34e-3, MAE = 7.48e-4, RMSE = 9.74e-4), demonstrating superior accuracy. RIFNO also performs well with slightly higher errors (\(\mathcal{R}\) = 1.63e-3, MAE = 8.79e-4, RMSE = 1.17e-3). DeepONet and WNO also capture the hysteresis well but with an order of magnitude less accuracy than FNO. In contrast, RNN, LSTM, GRU, and EDLSTM show higher errors and cannot capture the hysteresis and predict for novel input $H$ fields. Table~\ref{tab:forc} underscores the effectiveness of FNO in accurately predicting first-order reversal curves, followed by RIFNO, DeepONet, and WNO while traditional recurrent models underperform.

\subsection{Rate-independency}

\begin{table}[ht]
    \centering
    \caption{Errors for testing on different rates}
    \begin{tabular}{|l|l|c|c|c|c|}
        \hline
        Rate & Metric & DeepONet & FNO & WNO & RIFNO \\ \hline
        \multirow{3}{*}{$\times \frac{1}{100}$} & $\mathcal{R}$ & 1.34e+0 & 9.37e-2 & 5.26e+0 & \textbf{1.63e-3}  \\ \cline{2-6}
                                & MAE  & 7.72e-1 & 5.25e-2 & 6.55e-1 & \textbf{8.79e-4}  \\ \cline{2-6}
                                & RMSE & 9.71e-1 & 6.77e-2 & 9.98e-1 & \textbf{1.17e-3} \\ \hline
        \multirow{3}{*}{$\times \frac{1}{10}$} & $\mathcal{R}$ & 1.12e+0 & 8.42e-2 & 6.24e+0 & \textbf{1.63e-3} \\ \cline{2-6}
                                & MAE  & 6.37e-1 & 4.71e-2 & 5.74e-1 & \textbf{8.79e-4} \\ \cline{2-6}
                                & RMSE & 8.14e-1 & 6.08e-2 & 9.08e-1 & \textbf{1.17e-3} \\ \hline
        \multirow{3}{*}{$\times 10$} & $\mathcal{R}$ & 6.05e+0 & 9.05e-1 & 8.72e+0 & \textbf{1.63e-3} \\ \cline{2-6}
                                & MAE  & 3.71e+0 & 4.89e-1 & 3.27e+0 & \textbf{8.79e-4} \\ \cline{2-6}
                                & RMSE & 4.37e+0 & 6.54e-1 & 9.49e+0 & \textbf{1.17e-3} \\ \hline
        \multirow{3}{*}{$\times 100$} & $\mathcal{R}$ & 6.51e+0 & 2.40e+0 & 9.51e+0 & \textbf{1.63e-3} \\ \cline{2-6}
                                & MAE  & 4.20e+0 & 1.35e+0 & 4.12e+0 &  \textbf{8.79e-4} \\ \cline{2-6}
                                & RMSE & 4.70e+0 & 1.73e+0 & 5.33e+0 &  \textbf{1.17e-3} \\ \hline
    \end{tabular}
    \label{tab:rate}
\end{table}

\begin{figure*}[!t]
    \centering
    \begin{subfigure}[b]{0.15\textwidth}
        \includegraphics[width=\textwidth]{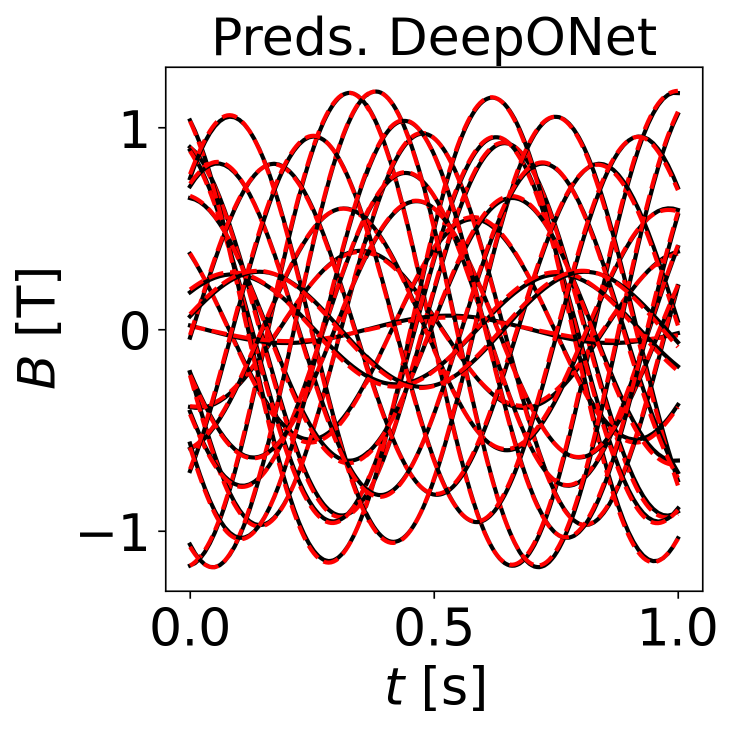}
    \end{subfigure}
    \begin{subfigure}[b]{0.15\textwidth}
        \includegraphics[width=\textwidth]{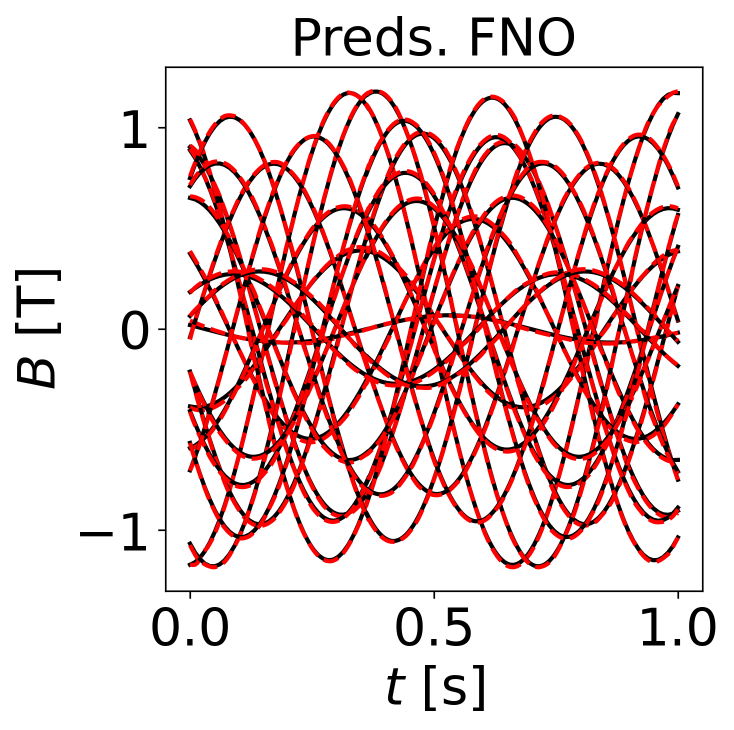}
    \end{subfigure}
    \begin{subfigure}[b]{0.15\textwidth}
        \includegraphics[width=\textwidth]{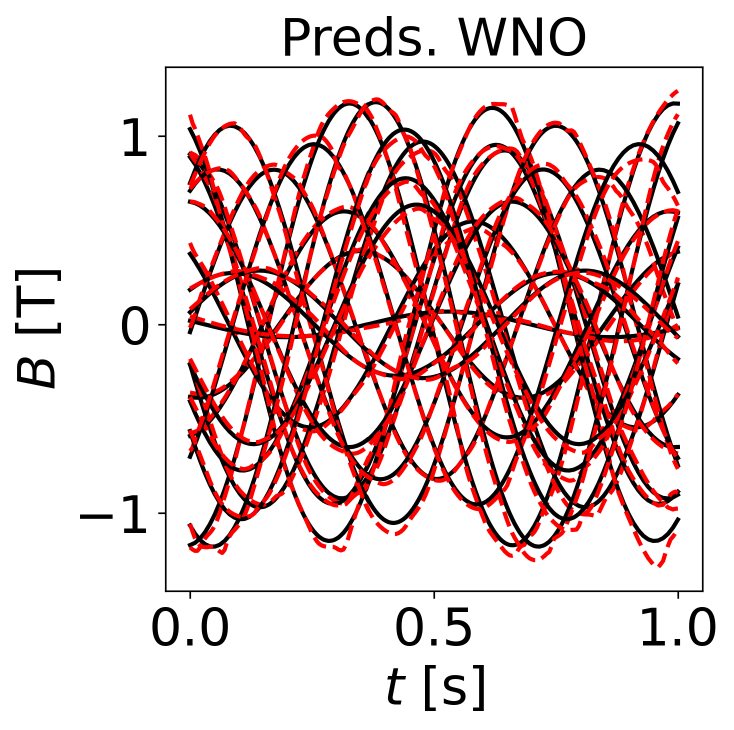}
    \end{subfigure}
    \begin{subfigure}[b]{0.15\textwidth}
        \includegraphics[width=\textwidth]{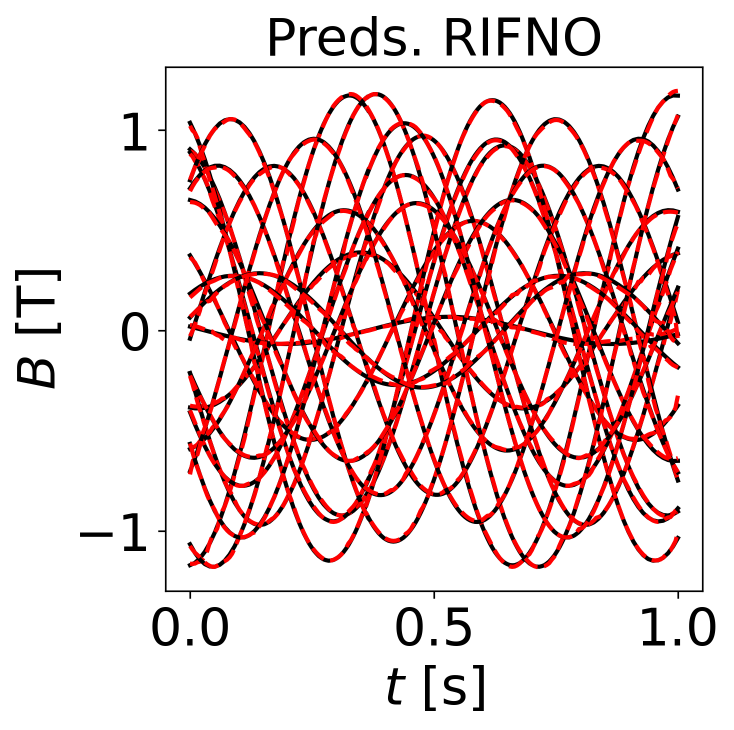}
    \end{subfigure}
    \begin{subfigure}[b]{0.15\textwidth}
        \includegraphics[width=\textwidth]{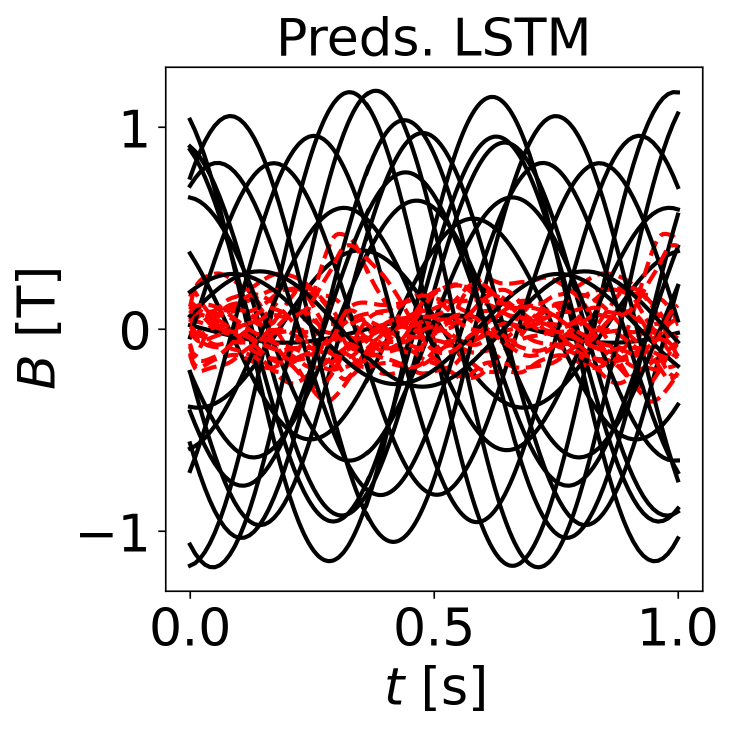}
    \end{subfigure}
    \begin{subfigure}[b]{0.15\textwidth}
        \includegraphics[width=\textwidth]{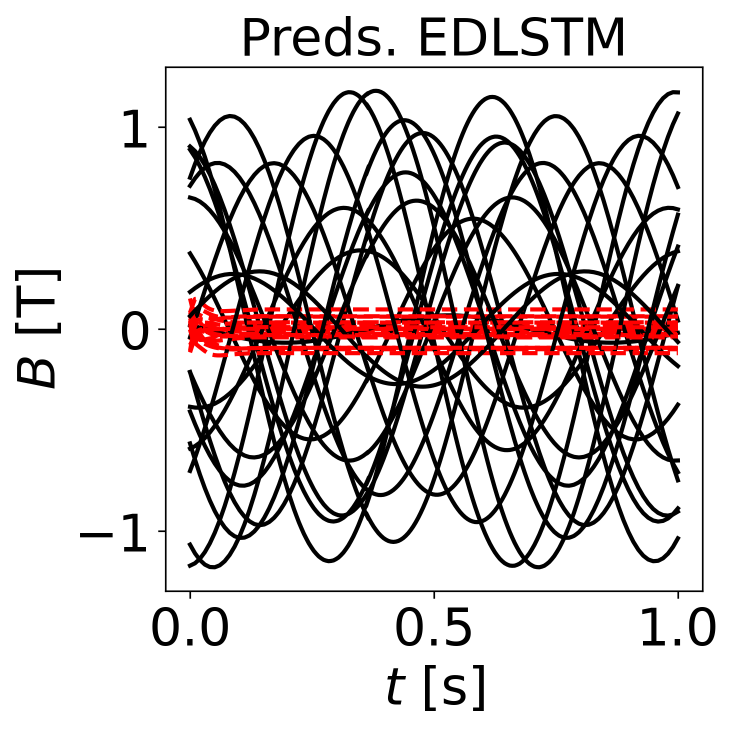}
    \end{subfigure}
    \vspace{0.25cm}

    \begin{subfigure}[b]{0.15\textwidth}
        \includegraphics[width=\textwidth]{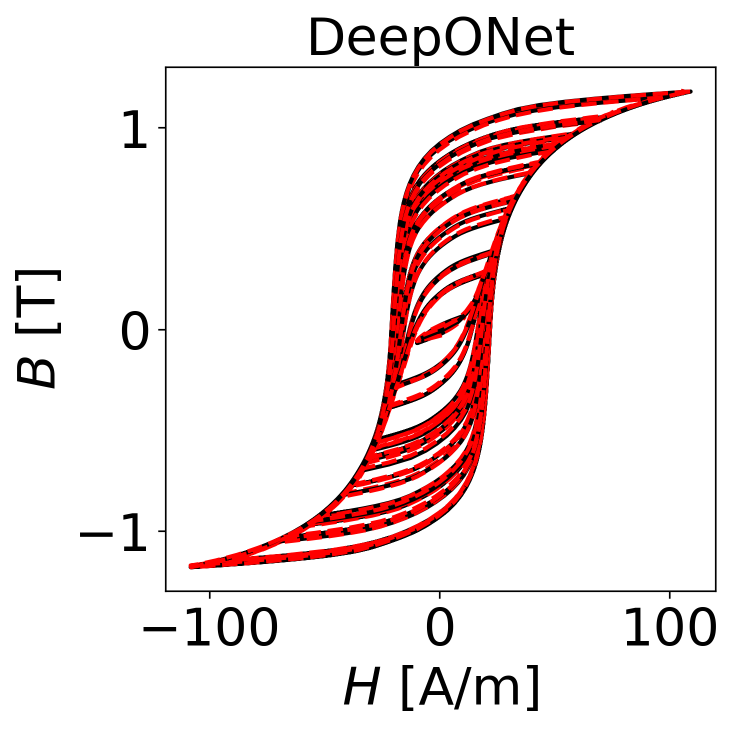}
    \end{subfigure}
    \begin{subfigure}[b]{0.15\textwidth}
        \includegraphics[width=\textwidth]{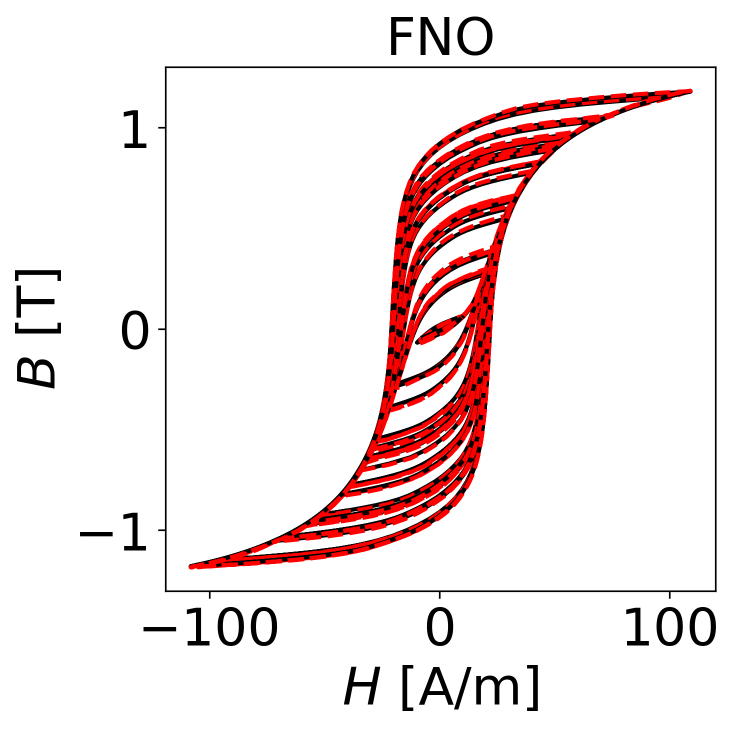}
    \end{subfigure}
    \begin{subfigure}[b]{0.15\textwidth}
        \includegraphics[width=\textwidth]{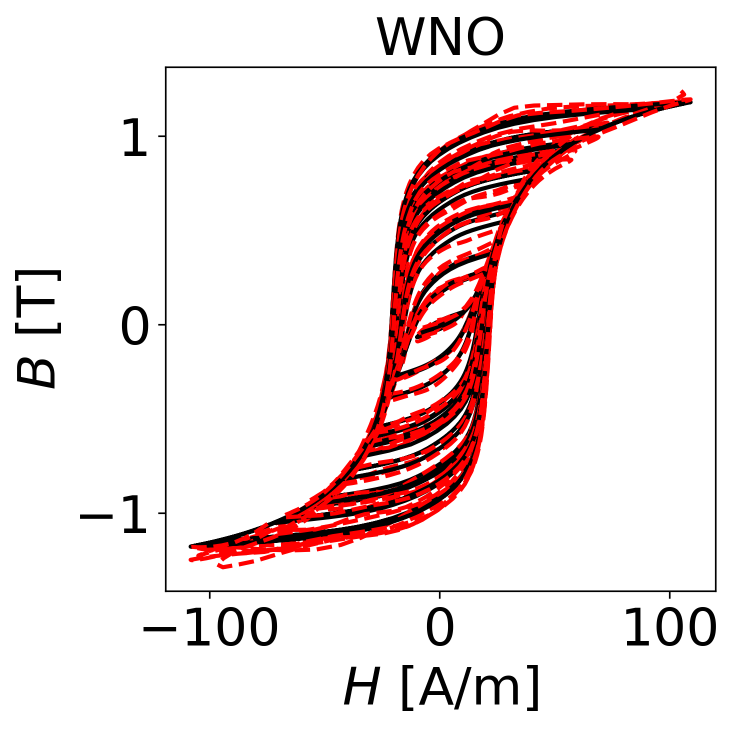}
    \end{subfigure}
    \begin{subfigure}[b]{0.15\textwidth}
        \includegraphics[width=\textwidth]{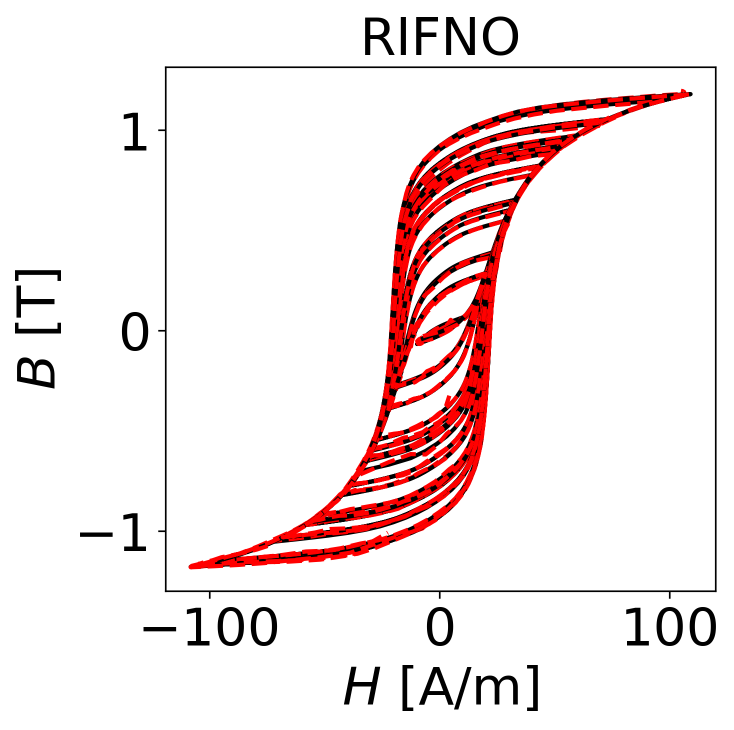}
    \end{subfigure}
    \begin{subfigure}[b]{0.15\textwidth}
        \includegraphics[width=\textwidth]{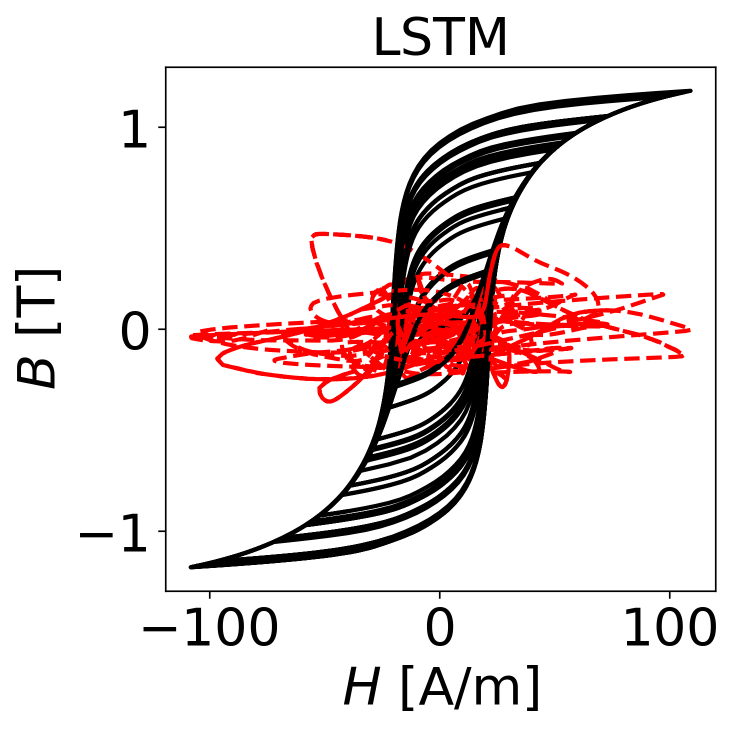}
    \end{subfigure}
    \begin{subfigure}[b]{0.15\textwidth}
        \includegraphics[width=\textwidth]{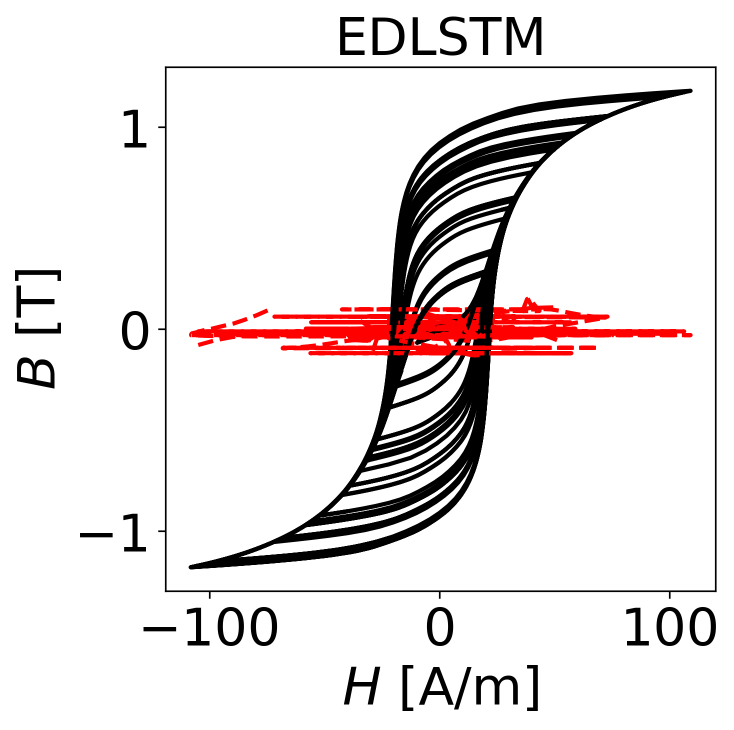}
    \end{subfigure}
    \vspace{0.25cm}

    \begin{subfigure}[b]{0.15\textwidth}
        \includegraphics[width=\textwidth]{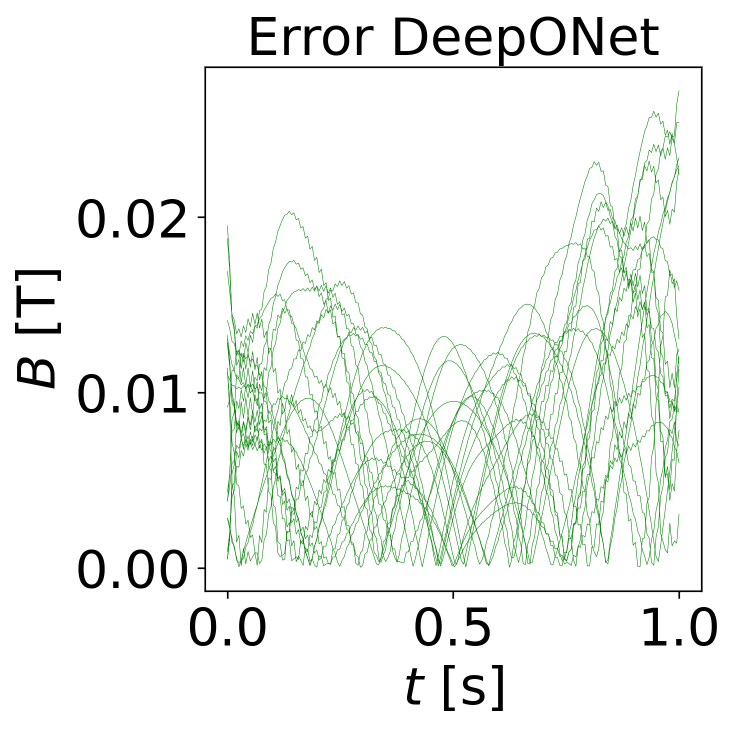}
    \end{subfigure}
    \begin{subfigure}[b]{0.15\textwidth}
        \includegraphics[width=\textwidth]{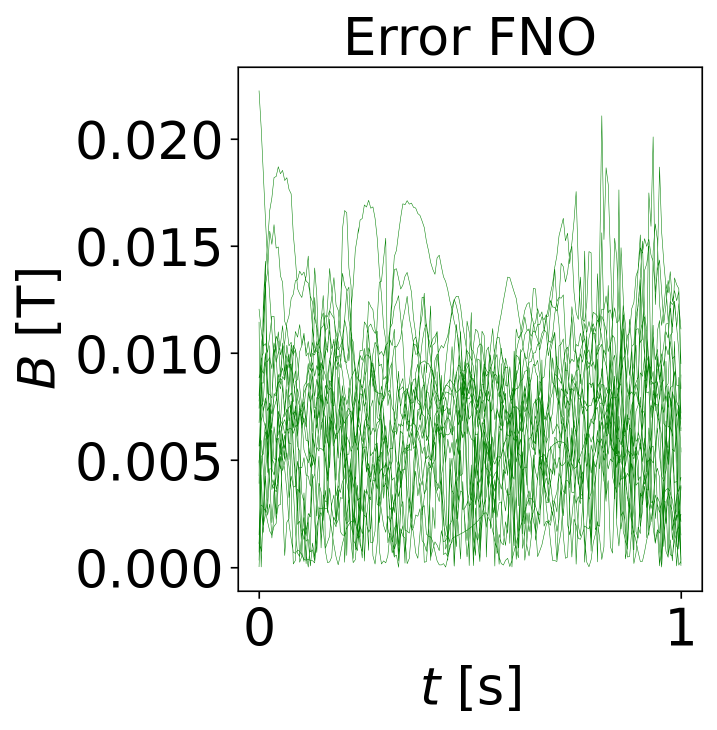}
    \end{subfigure}
    \begin{subfigure}[b]{0.15\textwidth}
        \includegraphics[width=\textwidth]{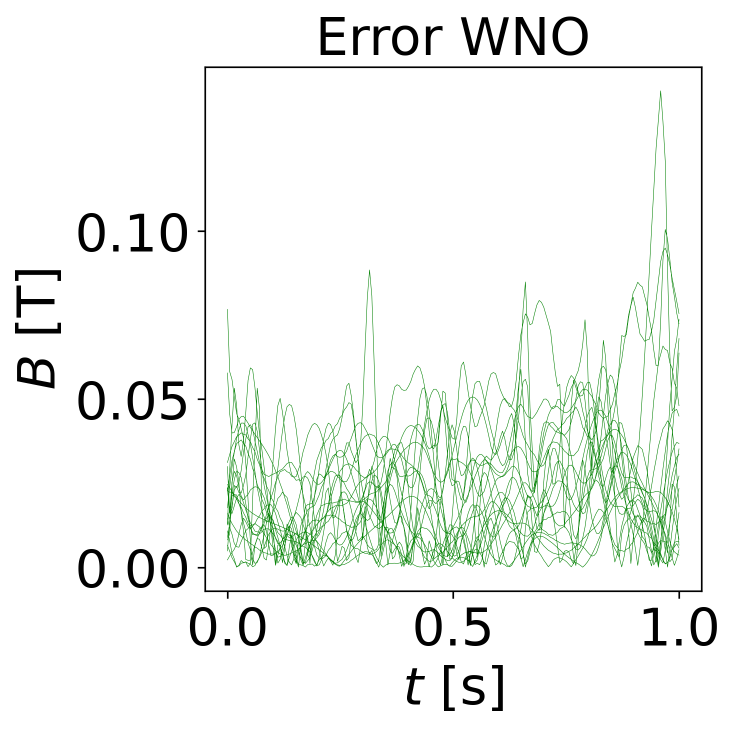}
    \end{subfigure}
    \begin{subfigure}[b]{0.15\textwidth}
        \includegraphics[width=\textwidth]{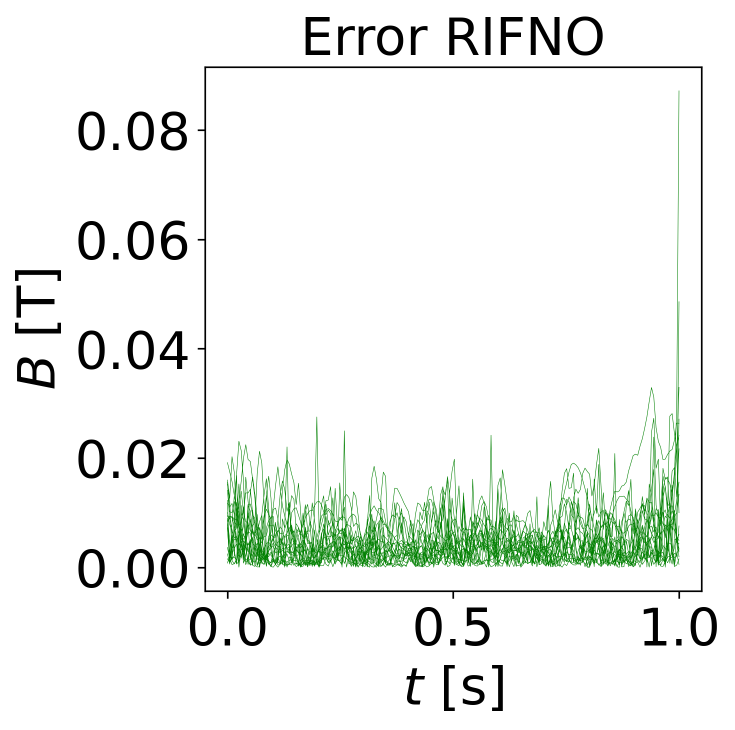}
    \end{subfigure}
    \begin{subfigure}[b]{0.15\textwidth}
        \includegraphics[width=\textwidth]{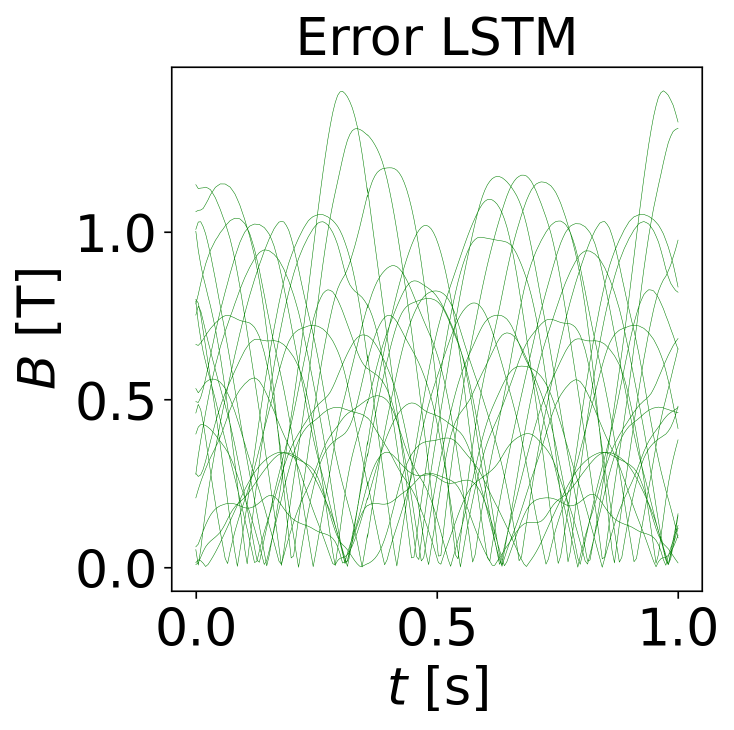}
    \end{subfigure}
    \begin{subfigure}[b]{0.15\textwidth}
        \includegraphics[width=\textwidth]{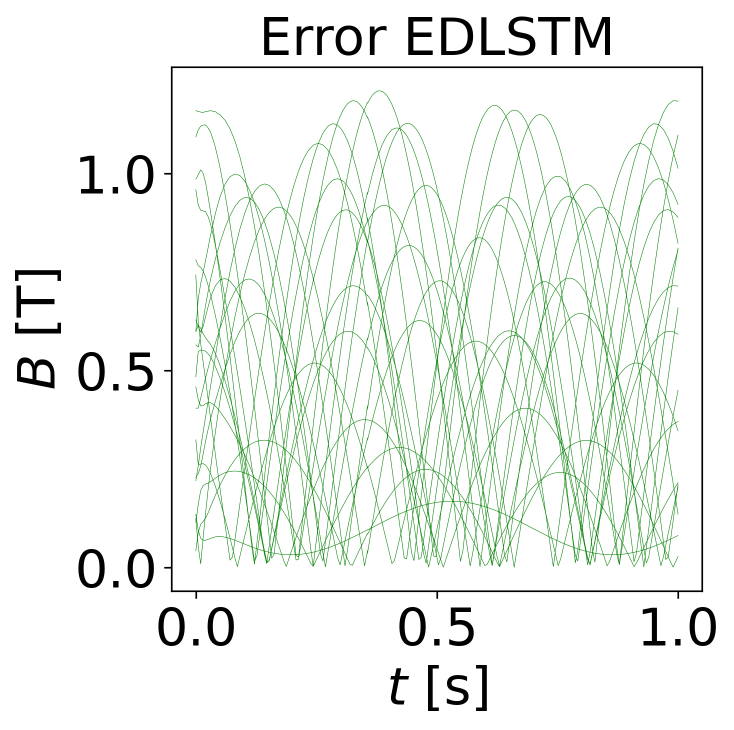}
    \end{subfigure}
    \caption{Performance of neural networks on predicting the minor loops. \textbf{Top row:} (From Left) the predictions (preds.) of the $B$ fields by DeepONet, FNO, WNO, RIFNO, LSTM, and Encoder-Decoder LSTM (EDLSTM) respectively. \textbf{Middle row:} (From Left) the predicted hysteresis loops (in red) compared with the reference (ref.) loops (in black) for the six methods. \textbf{Bottom row:} (From Left) absolute errors in predicting the minor loops for the six methods (in green).}
    \label{fig:minor}
\end{figure*}

This experiment evaluates the predictive capability of the trained FORC models under varying sampling rates during testing, differing from the training rate. Specifically, the original sampling array, $[0, 1]$, consisting of equispaced 198 points, is now tested using arrays $[0, 0.01]$, $[0, 0.1]$, $[0, 10]$, and $[0, 100]$ each at equispaced 198 points, corresponding to sampling rates of 1/100, 1/10, 10, and 100 times the rate used in training. This investigation aims to assess the model's ability to generalize across different sampling rates and determine if it maintains the rate-independent property of magnetic hysteresis.

Fig.~\ref{fig:line_curve} shows the error metric \(\mathcal{R}\) for DeepONet, FNO, RIFNO, and WNO across different testing rates compared to the training rate. The line graph demonstrates the variation of error for each model concerning the testing rate from \(\frac{1}{100}\) to \(100\) times the training rate. RIFNO exhibits the lowest and most consistent errors across all testing rates, indicating its superior generalization capability. FNO has higher errors compared to RIFNO. DeepONet, and WNO show the highest errors, especially at extreme testing rates, highlighting their comparatively lower robustness in varying conditions. Fig.~\ref{fig:line_curve} underscores the efficacy of RIFNO in maintaining low prediction errors across various testing scenarios, making it the most accurate operator among the three where different testing rates are considered.

Table~\ref{tab:rate} presents the errors in predicting the first-order reversal curve for different testing rates compared to the training rate using the DeepONet, FNO, RIFNO, and WNO models. The testing rates vary as \(\frac{1}{100}\), \(\frac{1}{10}\), \(10\), and \(100\) times the training rate, and the table includes three error metrics. For the \(\frac{1}{100}\) rate, RIFNO demonstrates the lowest errors with \(\mathcal{R}\) = 1.63e-3, MAE = 8.79e-4, and RMSE = 1.17e-3, outperforming DeepONet, FNO, and WNO. This trend continues across all testing rates, with RIFNO consistently showing minimal errors, indicating its robustness and superior accuracy. In contrast, DeepONet, and WNO exhibiting the highest errors, especially at extreme testing rates. FNO performs better than DeepONet, and WNO but still shows increasing errors as the testing rate deviates from the training rate.

\subsection{Minor loops}

\begin{table}[ht]
    \centering
    \caption{Errors in predicting minor loops}
    \begin{tabular}{|l|c|c|c|}
        \hline
        Method   & $\mathcal{R}$ & MAE & RMSE \\ \hline
        DeepONet & 1.80e-2 & 8.40e-3 & 1.00e-2  \\ \hline
        FNO      & 1.44e-2 & 7.08e-3 & 8.07e-3  \\ \hline
        WNO      & 5.92e-2 & 2.28e-2 & 2.95e-2  \\ \hline
        RIFNO    & \textbf{1.26e-2} & \textbf{5.07e-3} & \textbf{7.03e-3}   \\ \hline
        RNN      & 1.15e+0 & 5.25e-1 & 6.43e-1  \\ \hline
        LSTM     & 1.02e+0 & 4.74e-1 & 5.71e-1  \\ \hline
        GRU      & 1.06e+0 & 4.91e-1 & 5.95e-1  \\ \hline
        EDLSTM   & 1.01e+0 & 4.68e-1 & 5.61e-1  \\ \hline
    \end{tabular}
    \label{tab:minor}
\end{table}

\begin{table*}[ht]
    \centering
    \caption{Comparison of recurrent and operator architectures based on various attributes}
    \begin{tabular}{|l|c|c|c|c|c|c|c|c|}
        \hline
        \multirow{2}{*}{Attributes} & \multicolumn{8}{c|}{Methods} \\ \cline{2-9} 
                                    & RNN & LSTM & GRU & EDLSTM & DeepONet & FNO & WNO & RIFNO \\ \hline
        Predictions on trained excitation fields & \checkmark & \checkmark & \checkmark & \checkmark & \checkmark & \checkmark & \checkmark & \checkmark \\ \hline
        Preserves magnetization-demagnetization history    & \checkmark & \checkmark & \checkmark & \checkmark & \checkmark & \checkmark & \checkmark & \checkmark \\ \hline
        Maps functions to functions        & $\times$ & $\times$ & $\times$ & $\times$ & \checkmark & \checkmark & \checkmark & \checkmark \\ \hline
        Predictions on novel excitation fields & $\times$ & $\times$ & $\times$ & $\times$ & \checkmark & \checkmark & \checkmark & \checkmark \\ \hline
        Rate-independent                   & \checkmark & \checkmark & \checkmark & \checkmark & $\times$ & $\times$ & $\times$ & \checkmark \\ \hline
    \end{tabular}
    \label{tab:attributes}
\end{table*}

In the final numerical experiment, minor loops are predicted. The dataset for training and testing neural networks consists of functions generated using the cosine kernel of Gaussian processes. The input domain is $[0, 3\pi]$, with 198 evenly spaced points. These points are used with a cosine kernel to define covariance functions from which function samples are drawn. Each sample adheres to a maximum threshold of \SI{1.2}{\tesla}. A total of 2000 sine curves are generated and evenly split between training and testing sets, ensuring $N_{\text{train}} + N_{\text{test}} = 2000$. Consistent network hyperparameters are applied across all methods to maintain alignment with FORC prediction cases.

Fig.~\ref{fig:minor} illustrates the performance of various neural operator methods in predicting the minor loops. The first row shows predictions from DeepONet, FNO, WNO, RIFNO, LSTM, and EDLSTM with RIFNO providing the most accurate approximations. The second row presents the minor loop curves, comparing predicted (red curves) and actual (black curves) output fields, demonstrating that FNO closely matches the ground truth. The third row displays absolute error plots, showing that LSTM, and EDLSTM perform poorly with higher errors. The errors for DeepONet follow a specific pattern, while the errors for FNO predictions are minimal, with a maximum of approximately 0.02, whereas WNO predictions have a maximum absolute error around 0.01. The errors for RIFNO are prominent near the ends. 

Table~\ref{tab:minor} compares prediction errors for minor loops across eight models: DeepONet, FNO, WNO RIFNO, RNN, LSTM, GRU, and EDLSTM using relative error (\(\mathcal{R}\)), MAE, and RMSE. RIFNO achieves the lowest errors (\(\mathcal{R}\) = 1.26e-2, MAE = 5.07e-3, RMSE = 7.03e-3), indicating superior accuracy. FNO and DeepONet follow with slightly higher errors than RIFNO. WNO exhibits the maximum error amongst the neural operator methods. Conversely, traditional recurrent models like RNN, LSTM, GRU and recent methods like EDLSTM show significantly higher errors, with RNN performing the worst. Table~\ref{tab:minor} underscores the leading performance of RIFNO, followed by FNO, DeepONet, and WNO while traditional recurrent models lag in modeling novel minor loops not used for training the neural network model.

\section{Discussion} \label{sec:discussion}

This section discusses key implications following the successful application of neural operators in modeling hysteresis. The presented numerical experiments on FORCs and minor loops demonstrate that neural operators accurately predict the magnetic field response \( b \sim B \) corresponding to \( h \sim H \), whether \( h \) is part of the training data or not. In contrast, recurrent architectures only reliably predict responses for \( h \) values within the training set, highlighting a limitation for applications requiring predictions for novel magnetic fields. Moreover, once trained, neural operators enable real-time inference, enhancing their utility in applications such as control.

A comparative analysis between recurrent and operator-based methods is presented in Table~\ref{tab:attributes}, focusing on essential attributes for modeling magnetic hysteresis. Table~\ref{tab:attributes} illustrates that operator learning methods excel in mapping arbitrary \( H \) fields to \( B \) fields while preserving the history of magnetization and demagnetization. The recurrent networks are rate-independent owing to their architecture, which does not explicitly include the rate at which the data is collected. RIFNO provides an advantage over recurrent and other operator networks by offering function-to-function mapping and rate independence properties.

Fundamentally, the results bridge the gap between hysteresis modeling and SciML. The evolution of SciML has led to the development of a plethora of operator learning methods to accelerate the modeling and simulation of physical systems, including but not limited to \cite{yang2023context, raonic2024convolutional, fanaskov2023spectral}. These operator learning methods could be employed to model hysteresis and advance modeling of magnetic materials. The bridge between hysteresis modeling and SciML also presents an avenue to benchmark the results and datasets as advocated in \cite{li2023magnetdata}. 

Future research directions include employing and developing different operator training methods. Techniques like transfer learning \cite{zhuang2020comprehensive}, leveraging pre-trained network parameters \cite{kapoor2024transfer}, and meta-learning approaches \cite{hospedales2021meta} can be utilized to accelerate model convergence. Additionally, the results presented herein for scalar hysteresis could be expanded upon to model vector hysteresis, leveraging the property that deep learning often circumvents the curse of dimensionality and generalizes well to higher dimensions. Furthermore, validating neural operators for experimental data is part of our future work, where the study will be extended to include experimental data for different magnetic materials.

\section{Conclusions} \label{sec:conclusions}
This paper introduced a novel approach to model magnetic hysteresis through neural operators. The research highlights the benefits of employing neural operators for hysteresis modeling, yielding generalizable models essential for scenarios where prior training on varying magnetic fields is impractical. By employing prominent operator networks, namely deep operator network, Fourier neural operator, and wavelet neural operator, the research demonstrated the prediction of novel first-order reversal curves and minor loops outperforming the traditional recurrent architectures and leading to the development of generalizable neural hysteresis models. Furthermore, this contribution introduced a rate-independent Fourier neural operator, which enhanced the neural operator framework by incorporating the rate-independent hysteresis characteristics prevalent in magnetic materials. Concretely, this paper advances neural hysteresis modeling by developing accurate and generalizable models capable of real-time inference of material responses.

\bibliographystyle{IEEEtran}
\bibliography{ref}

\end{document}